\documentclass[10pt,twocolumn,letterpaper]{article}

\usepackage[pagenumbers]{cvpr} 

\definecolor{cvprblue}{rgb}{0.21,0.49,0.74}
\usepackage[pagebackref,breaklinks,colorlinks,allcolors=cvprblue]{hyperref}

\def\paperID{8512} 
\def\confName{CVPR}
\def\confYear{2026}

\title{TReFT: Taming Rectified Flow Models For One-Step Image Translation}

\author{Shengqian Li$^{1,2}$\thanks{Equal contribution.}, Ming Gao$^{5,*}$, Yi Liu$^{3}$, Zuzeng Lin$^{4}$, Feng Wang$^{5}$, Feng Dai$^{2}$\thanks{Corresponding author.}\\
\textsuperscript{1}University of Chinese Academy of Sciences, Beijing, China\\ 
\textsuperscript{2}Institute of Computing Technology, Chinese Academy of Sciences, Beijing, China\\ 
\textsuperscript{3}Beihang University, Beijing, China \textsuperscript{4}Tianjin University, Tianjin, China \textsuperscript{5}CreateAI, Beijing, China \\
\footnotesize
\texttt{\{lishengqian24s, fdai\}}@ict.ac.cn, \texttt{\{dujinshidai, feng.wff\}}@gmail.com, \texttt{18373214}@buaa.edu.cn, \texttt{linzuzeng}@tju.edu.cn
}

\usepackage{multirow}
\usepackage{mdframed}

\begin{document}

\maketitle
\begin{abstract}
Rectified Flow (RF) models have advanced high-quality image and video synthesis via optimal transport theory. However, when applied to image-to-image translation, they still depend on costly multi-step denoising, hindering real-time applications.
Although the recent adversarial training paradigm, CycleGAN-Turbo, works in pretrained diffusion models for one-step image translation, we find that directly applying it to RF models leads to severe convergence issues.
In this paper, we analyze these challenges and propose {\bf TReFT}, a novel method to {\bf T}ame {\bf Re}ctified {\bf F}low models for one-step image {\bf T}ranslation. 
Unlike previous works, TReFT directly uses the velocity predicted by pretrained DiT or UNet as output—a simple yet effective design that tackles the convergence issues under adversarial training with one-step inference.
This design is mainly motivated by a novel observation that, near the end of the denoising process, the velocity predicted by pretrained RF models converges to the final clean image, a property we further justify through theoretical analysis.
When applying TReFT to large pretrained RF models such as SD3.5 and FLUX, we introduce memory-efficient latent cycle-consistency and identity losses during training, as well as lightweight architectural simplifications for faster inference.
Pretrained RF models finetuned with TReFT achieve performance comparable to sota methods across multiple image translation datasets while enabling real-time inference.
\end{abstract}    
\begin{figure}[htb]
  \centering
  \includegraphics[width=1\linewidth]{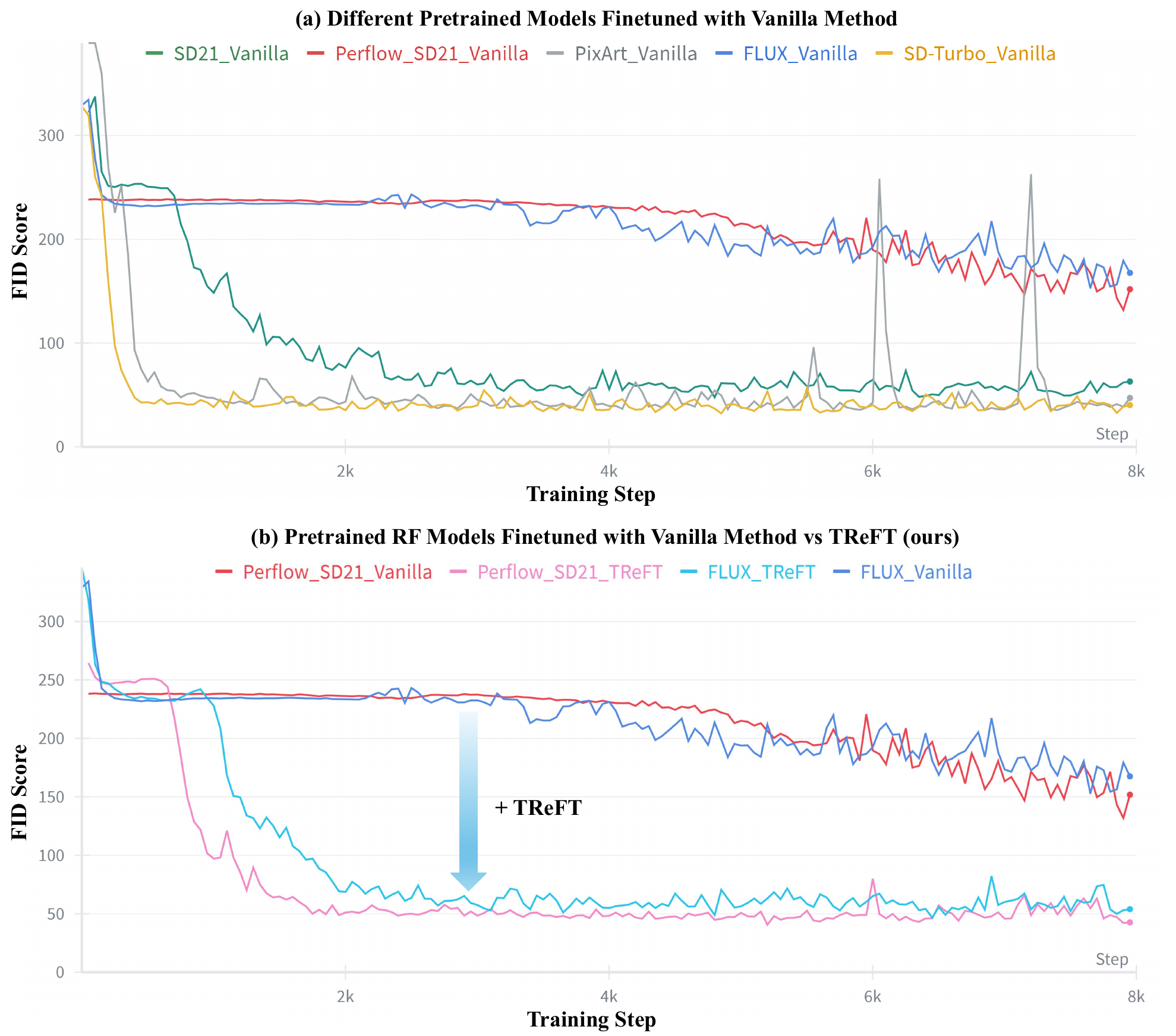}

   \caption{{\bf FID score calculated during training on Horse2zebra dataset.}
The experiment names indicate the pretrained models used: SD2.1\cite{rombach2022high}, PixArt\cite{chen2023pixart}, FLUX\cite{FLUX_website}, and SD-Turbo\cite{sauer2024adversarial}.
The suffixes “Vanilla” and “TReFT” denote the applied finetuning strategies, while the prefix “PerFlow” means the model is first finetuned using PerFlow~\cite{yan2024perflow}.
Please zoom in for details. See Appendix Sec.~\ref{sec:sup_exp_fig_1} for detailed experimental implementation.
}
   
   \label{fig:figure-1}
   \vspace{-1.2em}
\end{figure}
\section{Introduction}
\label{sec:intro}

Recent advances in Rectified Flow (RF) models \cite{rectflow, flowmatching, liu2023instaflow, albergo2022building} have enabled high-quality, text-conditioned image and video synthesis \cite{esser2024scaling, FLUX_website, Open-Sora} by leveraging optimal transport theory for faster sampling and denoising.
In image-to-image translation, RF models can incorporate the input image as a content condition via ControlNet \cite{controlnet} or through inversion-based paradigms \cite{meng2021sdedit, kulikov2024flowedit, wang2024taming, guo2024smooth}.
However, they still rely on multi-step denoising to achieve high-quality results, which incurs significant computational cost and limits their applicability in real-time scenarios.

Adapting large pretrained text-to-image models with adversarial objectives \cite{goodfellow2020generative} for one-step translation \cite{img2img-turbo} has shown promise on diffusion-based models like SD-Turbo \cite{sauer2024adversarial}.
However, when applied to RF models, this approach struggles to converge—even though RF and diffusion models share similar generation processes \cite{rectflow}.
Previous work has not investigated the cause of this discrepancy.

The key differences among these pretrained models lie in their backbones (UNet vs. DiT) and training objectives (Diffusion vs. Rectified Flow).
To investigate the underlying cause of the convergence issue, we conducted ablation experiments on the Horse→Zebra. Specifically, we compared SD-Turbo\cite{sauer2024adversarial} and PixArt-Alpha\cite{chen2023pixart} (which differ in backbone), as well as SD2.1\cite{rombach2022high} and its PeRFlow-finetuned variant \cite{yan2024perflow} (which differ in training objective), using Vanilla finetuning method as in CycleGAN-Turbo\cite{img2img-turbo}.

As shown in Fig.~\ref{fig:figure-1}~(a), PixArt (with a DiT backbone) using Vanilla quickly achieves performance comparable to that of SD-Turbo (with a UNet backbone), suggesting that the backbone difference is not the primary cause of the issue.
Similarly, SD2.1 with Vanilla fine-tuning also converges well. However, its PeRFlow-finetuned version fails to converge under the same Vanilla setting and exhibits a similar FID curve to that of FLUX with Vanilla.
This suggests that the convergence issue arises from the difference in training objectives.

\begin{figure}[t]
  \centering
  \includegraphics[width=1\linewidth]{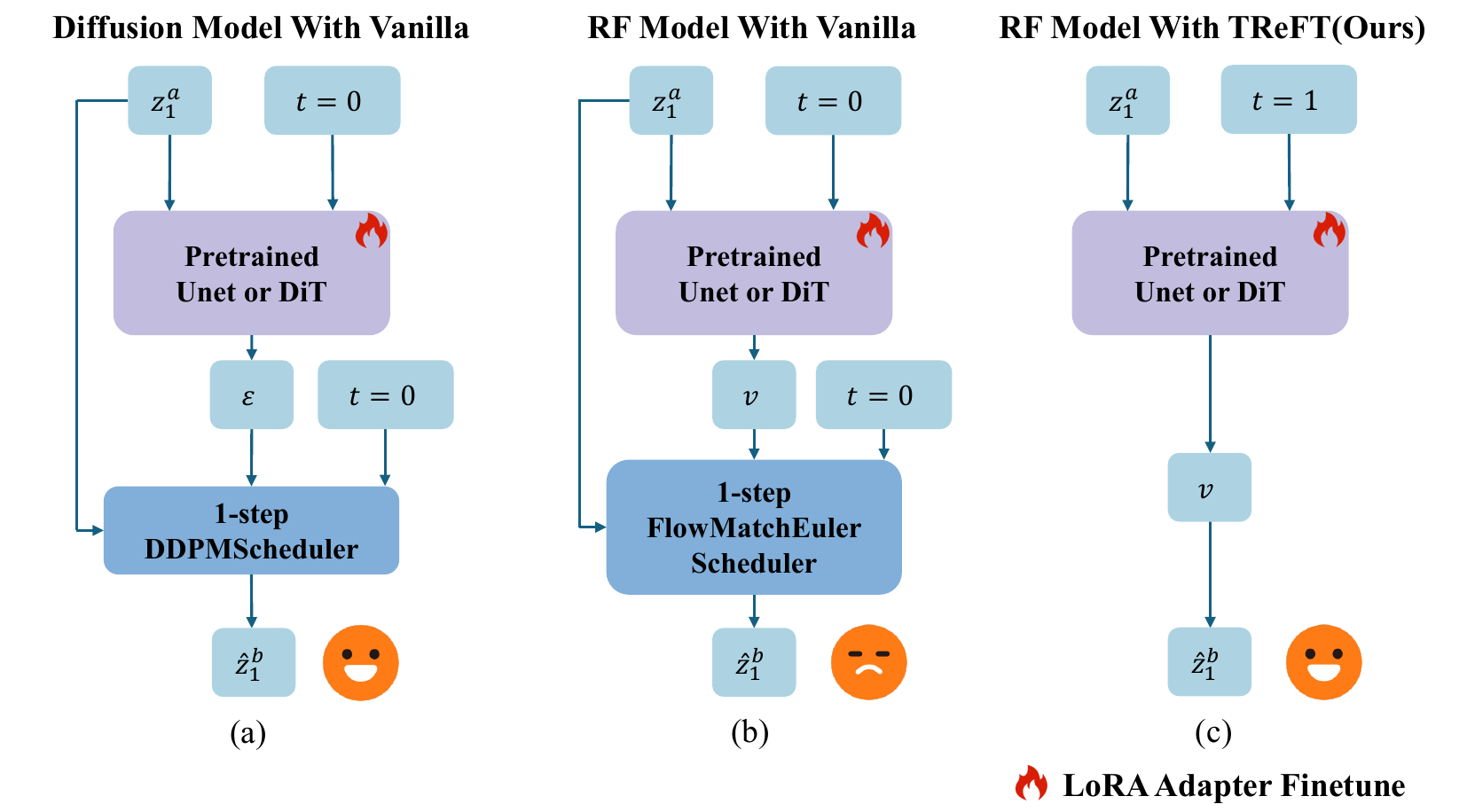}

   \caption{
    {\bf Comparison Between TReFT and Previous Paradigms.}
   (a) Diffusion models using the Vanilla method (e.g., CycleGAN-Turbo\cite{img2img-turbo}) take $z_1^a$ and timestep $t=0$ as input and output the one-step denoised image $\hat{z}_1^b$.
   (b) RF models using the Vanilla method.
   (c) RF models with TReFT take $z_1^a$ and timestep $t=1$ as input, and directly treat the prediction $v$ as the output $\hat{z}_1^b$. Happy: Easy to converge. Sad: Difficult to converge.
   Note: For simplicity, timesteps are unified. Here, $t=0$ is the state of pure noise, while $t=1$ corresponds to the clean image without noise.
   }
   
   \label{fig:figure-2}
   \vspace{-1.3em}
\end{figure}

In this paper, we propose {\bf TReFT}, a novel method to {\bf T}ame {\bf Re}ctified {\bf F}low models for one-step image {\bf T}ranslation.
The proposal of TReFT is motivated by two key observations.
First, by analyzing the trajectories in the VAE latent space produced by RF models fine-tuned using the Vanilla method, we observe that the pretrained RF model initially predicts a velocity pointing from noise toward the clean image, which aligns with the rectified flow theory. However, this differs significantly from the goal of image-to-image translation, which is to learn a velocity field between different image domains.
Second, visualization experiments on the predicted velocity of the pretrained RF model reveal that, during a standard multi-step generation process, the predicted velocity closely approximates the final clean image as the timestep approaches 1, where $t=1$ corresponds to a clean, noise-free image. We theoretically justify this phenomenon: under a Gaussian assumption on the latent distribution of prompt-conditioned images, we derive a closed-form characterization of the optimal velocity across all timesteps; moreover, under much weaker local smoothness assumptions, we prove that the predicted velocity still converges to the final clean image feature as the denoising process approches the end.

Based on these insights, TReFT directly uses the velocity predicted by pretrained DiT or UNet as output, which closely approximates the input clean image.
As Fig.~\ref{fig:figure-2} (c) displays, RF models with TReFT take $z_1^a$ and timestep $t=1$ as input, and directly treat the prediction $v$ as the output $\hat{z}_1^b$.
As Fig.~\ref{fig:figure-1} (b) displays, this simple yet effective design address convergence issues under adversarial training with one-step inference.
To reduce memory consumption during training and improve inference speed in implementation, we introduced two engineering optimizations: (1) latent cycle-consistency and identical losses; and (2) removal of text branches from the early MM-DiT blocks.

In summary, our core contributions are as follows:


\begin{itemize}
\item Through comparative experiments involving different backbones and objectives, we identified that the key reason for the convergence issues encountered when fine-tuning RF models using the Vanilla method on image translation datasets lies in the difference between the RF model’s objective and that of diffusion models;
\item We uncover a key property of pretrained RF models: near the end of denoising process, their predicted velocity converges to the clean image, and we provide theoretical justification for this behavior;
\item We propose TReFT, a simple yet effective approach for finetuning RF models that tackles convergence issues under adversarial training with one-step inference;
\item With our engineering optimizations, pretrained RF models finetuned with TReFT achieve performance comparable to sota methods while maintaining real-time inference speed across multiple image translation datasets.
\end{itemize}

\begin{figure*}[t]
    \centering
    \includegraphics[width=0.88\linewidth]{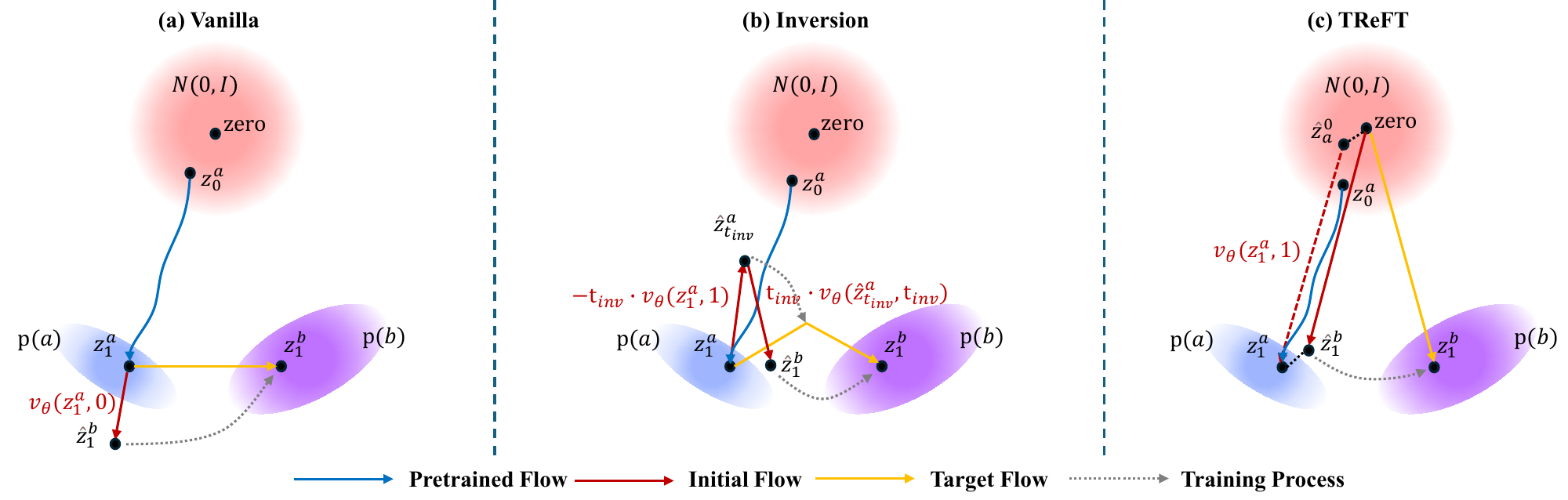}
    \caption{{\bf Pathways to Predict $\hat{z}_1^b$ in Latent Space for Three Methods: Vanilla, Inversion, and TReFT (Ours).}
    The four types of the lines represent: 
    pretrained flow (blue) from the pretrained RF model,
    initial flow (red) roughly aligned with its tangent direction,
    target flow (yellow) during training,
    and flow transition (grey) from initial to target.
    The three ellipse areas denote:
    noise distribution $N(0,I)$ (light red),
    source domain $p(a)$ (light blue),
    and target domain $p(b)$ (violet).
    The three methods illustrated are:
    (a) Vanilla: one-step denoising using standard rectified flow scheduler.
    (b) Inversion: one-step inversion followed by one-step denoising.
    (c) TReFT (ours): directly applies $v_{\theta}(z_1^a, 1)$ for translation.
     Note: To visualize $\hat{z}1^b$, $v{\theta}(z_1^a, 1)$ is shifted to start at the origin. This illustration is based on Sec.~\ref{sec:method-Preliminaries}, Sec.~\ref{sec:treft} and Fig.~\ref{fig:figure-10}.
    }


    \label{fig:figure-3}
\end{figure*}

\section{Related Work}
\label{sec:relate_work}

\subsection{Text-to-image models}

Diffusion models have rapidly emerged as a leading framework for high-quality image generation. Originally proposed by Sohl-Dickstein et al.~\cite{sohl2015deep}, they were later refined for image synthesis through denoising mechanisms~\cite{ho2020denoising, sde} and enhanced efficiency~\cite{ddpm, diffusion, song2023consistency, song2023improved, luo2023latent, rombach2022high}. Rectified Flow (RF)~\cite{rectflow, flowmatching} models generation process as an ODE between noise and data, offering faster sampling by learning velocity fields via optimal transport.
While many diffusion- and RF-based methods~\cite{meng2021sdedit, hertz2022prompt, parmar2023zero, tumanyan2023plug, wang2024instantstyle} enable zero-shot image editing and translation, our approach achieves one-step image translation with better performance and limited training expense.

\subsection{Image-to-image translation}

Image-to-image translation converts images from a source domain to a target domain. Depending on whether paired training data is available, methods are divided into supervised and unsupervised.

{\bf Supervised image-to-image translation} learns mappings using paired labeled data. Pix2Pix~\cite{pix2pix} first introduced conditional GANs combining adversarial and L1 losses. Pix2PixHD~\cite{pix2pixhd} improved this for high-resolution images using hierarchical generators and multi-scale discriminators. Later works added features like semantic normalization (SPADE~\cite{park2019semantic}), sketch-to-image translation (Scribbler~\cite{sangkloy2017scribbler}), and style-aware normalization (SEAN~\cite{zhu2020sean}). However, these methods still require large amounts of paired data, which is often hard to obtain in practice.

\textbf{Unsupervised Domain Translation} learns mappings between domains without paired data. Early works like CycleGAN~\cite{cyclegan}, DualGAN~\cite{yi2017dualgan}, and DiscoGAN~\cite{kim2017learning} introduced cycle consistency constraints. Later methods improved results with contrastive learning, better losses, and disentangled representations \cite{cut, han2021dual, shrivastava2017learning, taigman2016unsupervised, munit, lee2018diverse}. Recent diffusion-based approaches \cite{unit-ddpm, su2022dual, wu2023latent} boost fidelity via iterative denoising, and CycleGAN-Turbo \cite{img2img-turbo} enables one-step translation using SD-Turbo \cite{sauer2024adversarial}. Unlike these, our work uses pretrained RF models and addresses their convergence issues under the adversarial training paradigm.

\section{Method}
\label{sec:method}


\subsection{Preliminaries}
\label{sec:method-Preliminaries}




{\bf Rectified Flow} \cite{rectflow} accelerates the transition from the Gaussian noise distribution $\pi_0$ to the data distribution $\pi_1$ via a straight-line path. It models the transformation using an ordinary differential equation (ODE), also known as the rectified flow:
\begin{equation}
d z_t=v(z_t, t)d t, \quad where \, z_0 \sim \pi_0 \, and \, z_1 \sim \pi_1.
\end{equation}

The forward process is defined as a linear interpolation between $z_0$ and $z_1$:
$z_t = t z_1 + (1 - t) z_0$,
which implies the ODE:
$d z_t = (z_1 - z_0) d t$.
The model trains a neural network to approximate the velocity field $v_{\theta}(z_t, t)$ by minimizing the following regression loss:
\begin{equation}
\min _{\theta} {E} \Bigg[ \int_0^1 \left\|\left(z_{1}-z_{0}\right)-v_{\theta}\left(z_{t}, t\right)\right\|^{2} d t \Bigg].
\label{eq:equation-2}
\end{equation}

During sampling, the ODE is discretized using the Euler method. Given a sequence of $N$ timesteps ${t_N, ..., t_0}$, the process starts from noise $z_{t_N} \sim \mathcal{N}(0, I)$ and updates iteratively as:
\begin{equation}
z_{t_{i-1}}=z_{t_{i}} + (t_{i-1}-t_{i}) v_{\theta}\left(z_{t_{i}}, t_{i}\right).
\label{eq:equation-3}
\end{equation}
In text-to-image generation, a conditional velocity field $v_{\theta}(z_t, t, C)$ is learned, where $C$ represents the input text prompt.

\begin{figure}[ht]
    \centering
    \includegraphics[width=1\linewidth]{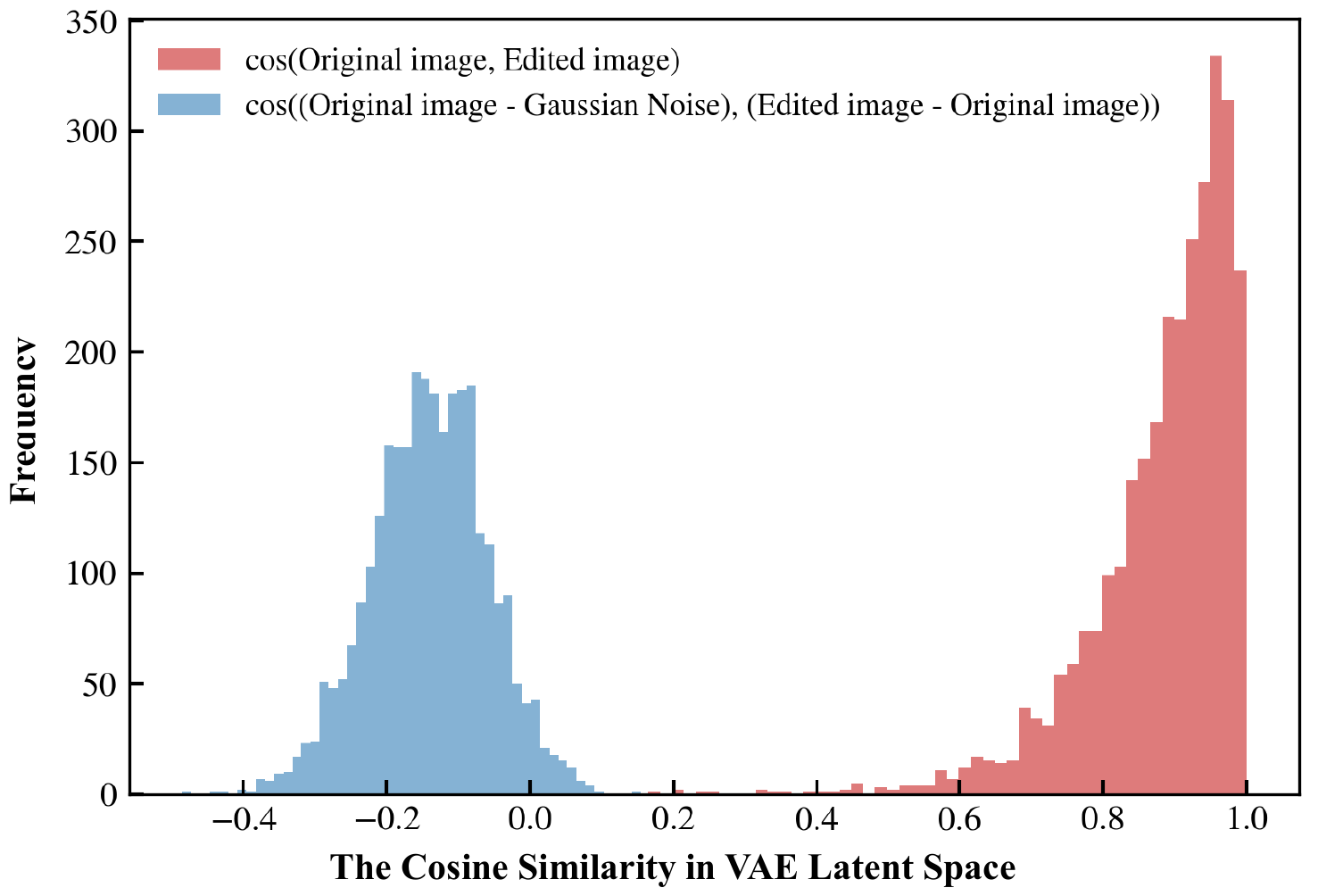}
    \caption{{\bf The Cosine Similarity in VAE Latent Space.} To evaluate the cosine similarity in VAE latent space, we conduct the experiment on 3,582 original–edited image pairs from the InstructPix2Pix CLIP-filtered Dataset\cite{brooks2023instructpix2pix, Instructpix2pix_Clip_Filtered_dataset} using the VAE of SD3.5-Large\cite{esser2024scaling}, and visualize the results as histograms. The blue histograms (Vanilla) indicate that the pretrained and target flows are nearly orthogonal, whereas the red histograms (TReFT) reveal that the directions of the original and edited image latents are closely aligned.
    }
    \label{fig:figure-10}
\end{figure}

\noindent{\bf One-step Image-to-image with pretrained RF models.}
Given an unpaired sample $z_1^a \sim p(a)$, we denote its ideal translation in the target domain $p(b)$ as $z_1^b$. We define image-to-image translation using pretrained rectified flow (RF) models as mapping the source-domain latent code $z_1^a$ to its target-domain counterpart $z_1^b$ (Fig.~\ref{fig:figure-3}).  
We introduce two common approaches below:

\textbf{\textit{Vanilla.}}(Fig.~\ref{fig:figure-2}~(b) and Fig.~\ref{fig:figure-3}~(a))
Following rectified flow denoising process~\cite{rectflow}, the pretrained RF model takes $z_1^a$ and timestep $t = 0$ as input, and performs one-step denoising to predict the target-domain image:
\begin{equation}
    \hat{z}_1^b=z_1^a + v_{\theta}\left(z_1^a, 0\right).
\end{equation}
An adversarial loss between the prediction $\hat{z}_1^b$ and ground truth $z_1^b$ encourages the initial flow $v_{\theta}(z_1^a, 0)$ to approximate the target displacement $z_1^b - z_1^a$.

\textbf{\textit{Inversion.}}(Fig.~\ref{fig:figure-3}~(b))
This method follows the standard inversion process in rectified flow models~\cite{meng2021sdedit, rectflow}.  
Let $t_{inv}$ be the backward step size. As shown in Fig.~\ref{fig:figure-3}(b), the process first applies a backward step from $z_1^a$ to $\hat{z}_{t_{inv}}^a$, then performs one-step denoising to obtain the predicted result:
\begin{equation}
    \hat{z}_1^b=z_1^a - t_{inv} v_{\theta}\left(z_1^a, 1\right) + t_{inv} v_{\theta}\left(\hat{z}_{t_{inv}}^a, t_{inv}\right).
\end{equation}
The combined flow $- t_{inv} v_{\theta}\left(z_1^a, 1\right) + t_{inv} v_{\theta}\left(\hat{z}_{t_{inv}}^a, t_{inv}\right)$
is trained via adversarial loss to approximate the target flow $z_1^b - z_1^a$. Both the inversion and denoising steps are jointly optimized, yielding the yellow flow path in Fig.~\ref{fig:figure-3}~(b).

\subsection{TReFT}
\label{sec:treft}



TReFT (Fig. \ref{fig:figure-3} (c)) is motivated by two novel observations.

\vspace{0.3em}
\noindent{\bf Observation 1.}
We observe that the discrepancy between the pretrained flow and the target flow imposed by the adversarial objective affects the convergence difficulty during finetuning.
By analyzing the experimental results shown in Fig.~\ref{fig:figure-1}~(a) and Fig.~\ref{fig:figure-10}, as well as the flow trajectories in the latent space illustrated in Fig.\ref{fig:figure-3} (a) and (b),
we find that for the Vanilla method, the direction of the pretrained flow is nearly orthogonal to that of the target flow (from Fig.~\ref{fig:figure-3}~(a) and blue histograms in Fig.~\ref{fig:figure-10}), making the training process difficult to converge.
In contrast, for the Inversion method, where both steps are trained simultaneously,
the gap between the pretrained flow and the target flow is smaller (as shown in Fig.\ref{fig:figure-3}),
which facilitates more stable and efficient convergence during training.
To address convergence issue, our TReFT method uses a target that is easier to optimize.


\vspace{0.3em}
\noindent{\bf Observation 2.}
During multi-step generation, the velocity predicted by the pretrained RF model gradually approaches the input image as the timestep $t$ increases. Although it aims to approximate $z_1 - z_0$, the noise component diminishes over time, and notably, as $t \to 1$, the predicted velocity $v_{\theta}(z_t, t)$ converges to the clean image $z_1$. This can be explained from the rectified flow training objective~\cite{rectflow, liu2022rectified}, where minimizing an $L_2$ loss yields the conditional expectation~\cite{bishop2006pattern}:
\begin{equation}
    v_{\theta}(z_t,t)=E \left[ z_1-z_0 \vert z_t,t \right].
    \label{eq:equation-5}
\end{equation}


\noindent{\bf Theorem 1.}
Let $z_0 \sim \mathcal N(0, I_d)$, and $z_1 \sim P(z_1 \mid c)$ the clean image latent conditioned on text prompt $c$.
Assume that $P(z_1 \mid c)$ can be approximated by a Gaussian $\mathcal N(\mu, \sigma^2 I_d)$.
Then, for the intermediate latent $z_t$, the closed-form conditional expectation of the flow-matching target is:

\begin{equation}
E[z_1 \!-\! z_0 | z_t,t] = \frac{t\sigma^2 \!-\! (1 \!-\! t)}{t^2\sigma^2 \!+\! (1 \!-\! t)^2} z_t + \frac{(1 \!-\! t)}{t^2\sigma^2 \!+\! (1 \!-\! t)^2} \mu.
\label{eq:equation-6}
\end{equation}

\noindent{\em Derivation outline of Theorem 1.}  This result can be derived from linear-Gaussian models\cite{murphy2012machine}: since $(z_1, z_t)$ and $(z_0, z_t)$ are jointly Gaussian, their conditional expectations admit a closed-form linear expression, whose difference yields Eq.~\ref{eq:equation-6}. Proof details are in Appendix Sec.~\ref{sec:sup_proof_theorem_1}.

Combining Eq.~\ref{eq:equation-5} and Eq.~\ref{eq:equation-6} gives the limiting behavior:
\begin{equation}
    \underset{t \to 1}{lim} \, v_{\theta}(z_t,t) = \lim_{t \to 1} E[z_1 - z_0 \mid z_t] = z_1.
    \label{eq:equation-7}
\end{equation}

\noindent{\bf Beyond the Gaussian assumption.}  
While the above derivation assumes a Gaussian form for $P(z_1 \mid c)$, this can be restrictive in practice due to varying text prompts.
In Theorem 1, we analyze the expected velocity for all timesteps $t \in (0,1)$, considering $z_1$ as a random latent variable sampled from the conditional distribution $P(z_1 \mid~c)$.  
In contrast, Theorem 2 focuses on the limiting behavior as $t \to~1$, where the conditional expectation converges to the final fixed latent feature $z_1^*$, representing the final clean image latent vector drawn from $P(z_1 \mid c)$.
This local limit result requires only mild $C^{1,1}$ \footnote{$C^{1,1}$ means continuously differentiable with a locally Lipschitz gradient\cite{evans2018measure}, implying bounded curvature without requiring second derivatives.} smoothness conditions near $z_1^*$, which is consistent with the locally smooth latent manifolds in VAE latent space\cite{vae,shao2018riemannian}.

\begin{figure}[tb]
    \centering
    \includegraphics[width=1\linewidth]{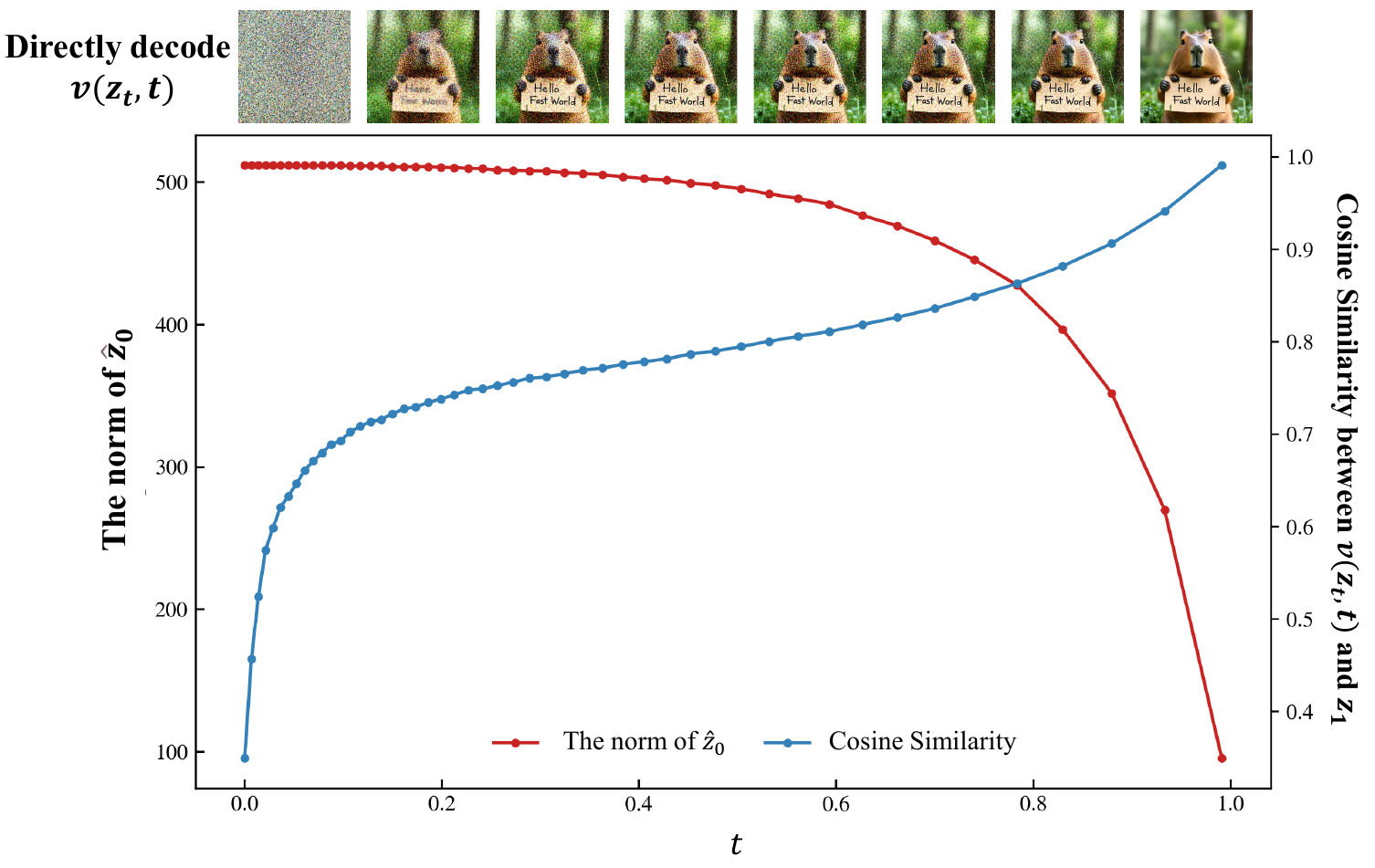}
  
     \caption{
\textbf{The norm of $\hat{z}_0$ and cosine similarity between $v(z_t, t)$ and $z_1$ at different timesteps, with the visualizations of $v(z_t, t)$.}
The image sequence above is generated directly by passing $v(z_t, t)$ through the VAE decoder.
In the lower plot picture, the red curve corresponds to the left vertical axis and represents the norm of $\hat{z}_0$ at each timestep.
The blue curve corresponds to the right vertical axis and shows cosine similarity between $v(z_t, t)$ and $z_1$ at different timesteps.
This experiment is conducted on SD3.5-Large\cite{esser2024scaling}, sampling 50 steps to generate $1024 \times 1024 \times 3$ images  on 1000 different prompts.}
    \label{fig:figure-4}
\end{figure}

\noindent{\bf Theorem 2.}
Let $z_0 \sim \mathcal N(0, I_d)$ and $z_1 \sim P(z_1 \mid c)$.
At the end of denoising process, assume that the conditional density $p(z_1 \mid c)$ is strictly positive and $C^{1,1}$ smooth in a neighborhood $U$ of the final latent feature $z_1^*\sim P(z_1 \mid c)$. The conditional expectation satisfies:

\begin{equation}
    \lim_{t \to 1} E[z_1 - z_0 \mid z_t] = \lim_{t \to 1}(z_1^* + O((1-t))) = z_1^*.
\end{equation}

\noindent{\em Derivation outline of Theorem 2.}  
The proof is based on a local Laplace expansion of the posterior $p(z_1 \mid z_t)$ around its mode $\hat z = z_t / t$, which converges to $z_1^*$ as $t \to 1$.
Under the $C^{1,1}$ smoothness assumption, the posterior mean asymptotically approaches $z_1^*$ with a bias of order $O(1-t)$, consistent with standard Laplace approximations~\cite{tierney1986accurate}.
Detailed derivations are provided in Appendix Sec.~\ref{sec:sup_proof_theorem_2}.



\noindent{\bf Experiment verification of Theorem 1 and 2.} To verify this, we designed an experiment simulating the image generation process of a pretrained RF model on a LLM-generated prompt dataset.
We measure cosine similarity between the predicted velocity and the final image at each timestep to evaluate their difference, and compute the norm of the noise vector to analyze noise components, which is predicted via one-step inversion:
\begin{equation}
    \hat{z}_0=z_t - t v_{\theta}\left(z_t, t\right).
    \label{eq:equation-8}
\end{equation}

As shown in Fig.~\ref{fig:figure-4}, based on SD3.5-Large, as $t$ approaches 1, the norm of the predicted noise $\hat{z}_0$ decreases and $v_{\theta}(z_t, t)$ approaches the final image.
This trend is consistent across all pretrained RF models, whether distilled or not.
See the Appendix Sec.~\ref{sec:sup_exp_theorem} for more experiment on other pretrained RF models.


  

\vspace{0.3em}
\noindent{\bf TReFT.}
Encouraged by the two observations above, we propose TReFT.
As shown in Fig.\ref{fig:figure-2}~(c) and Fig.\ref{fig:figure-3}~(c), the output of the generative model is remarkably simple—namely, we directly use the output of DiT as the predicted image in the target domain:
\begin{equation}
    \hat{z}_1^b=v_{\theta}\left(z_1^a, 1\right).
\end{equation}
This is a simple yet effective approach.
In the latent space, TReFT directly uses $v_{\theta}(z_1,1)$ as the prediction of $z_1^b$.
The adversarial learning objective applied to $\hat{z}_1^b$ and $z_1^b$ encourages the initial flow from $v_{\theta}(z_1,1)$ to $z_1^b$.
This is easy to learn because $z_1^b$ which can be viewed as $z_1^b-\mathbf{0}$, lies along the direction from noise to image.



\begin{table*}[t]
    \centering
    {
    \fontsize{9pt}{11pt}\selectfont
    \begin{tabular}{c|cccc|cccc}
    \toprule
      \multirow{3}{*}{Method} & \multicolumn{2}{l}{Horse $\rightarrow$ Zebra} & \multicolumn{2}{l}{Zebra $\rightarrow$ Horse} & \multicolumn{2}{l}{Day $\rightarrow$ Night} & \multicolumn{2}{l}{Night $\rightarrow$ Day} \\
                              & FID$\downarrow$ &\begin{tabular}[c]{@{}l@{}}DINO \\ Struct$\downarrow$\end{tabular} & FID$\downarrow$     &\begin{tabular}[c]{@{}l@{}}DINO \\ Struct$\downarrow$\end{tabular} & FID$\downarrow$  &\begin{tabular}[c]{@{}l@{}}DINO \\ Struct$\downarrow$\end{tabular} & FID$\downarrow$   &\begin{tabular}[c]{@{}l@{}}DINO \\ Struct$\downarrow$\end{tabular}  \\
      \midrule
      CycleGAN \cite{cyclegan}            & 74.9            & 3.2                    & 133.8               & 2.6                & 36.3                 & 3.6                 & 92.3                 & 4.9                 \\
      CUT \cite{cut}                      & 43.9            & 6.6                    & 186.7               & 2.5                & 40.7                 & 3.5                 & 98.5                 & 3.8                 \\
      \midrule
      SDEdit \cite{meng2021sdedit}       & 77.2            & 4.0                    & 198.5               & 4.6                & 111.7                & 3.4                 & 116.1                & 4.1                 \\
      Plug\&Play \cite{tumanyan2023plug} & 57.3            & 5.2                    & 152.4               & 3.8                & 80.8                 & {2.9}     & 121.3                & {\bf 2.8}           \\
      Pix2pix-Zero \cite{parmar2023zero} & 81.5            & 8.0                    & 147.4               & 7.8                & 81.3                 & 4.7                 & 188.6                & 5.8                 \\
      Cycle-Diffusion \cite{wu2023latent}& 38.6            & 6.0                    & 132.5               & 5.8                & 101.1                & 3.1                 & 110.7                & 3.7                 \\
      DDIB \cite{su2022dual}             & 44.4            & 13.1                   & 163.3               & 11.1               & 172.6                & 9.1                 & 190.5                & 7.8                 \\
      InstructPix2Pix\cite{brooks2023instructpix2pix} & 51.0 & 6.8  & 141.5 & 7.0  & 80.7 & {\bf 2.1} & 89.4 & 6.2 \\
      CycleGAN-Turbo \cite{img2img-turbo}& 41.0            & {\bf 2.1}              & {127.5}   & {1.8}          & {31.1}     & 3.0                 & {45.2}     & 3.8                 \\
      Flowedit \cite{kulikov2024flowedit}& {\bf 36.5}            & 9.5              & {146.9}   & {9.5}          & {112.0}     & 3.7                 & {131.4}     & \underline{3.6}                 \\
      \midrule
      SD3.5 w/TReFT (Ours)                 & \underline{38.5}& \underline{2.2}       & {\bf 119.6}         & \underline{2.2}    & \underline{30.9}            & {2.9}     & \underline{44.9}           & 3.7                 \\
    FLUX.1 w/TReFT (Ours)                 & {46.6}& {2.7}       & \underline{125.1}         & {\bf1.7}    & {\bf 29.9}            & \underline{2.8}     & {\bf 44.4}           & 3.7                 \\
      \bottomrule
    \end{tabular}
    }
    \caption{{\bf Comparison on unpaired datasets.} The best scores are marked in bold, and the second best scores are underlined.}
    \label{tab:tab-1}
\end{table*}

\vspace{0.3em}
\noindent{\bf Discussion.}
Returning to Fig.~\ref{fig:figure-3}, the three methods differ in convergence difficulty due to their distinct output forms under the same adversarial objective.
Vanilla struggles to converge, as it learns flow between real images, while the pretrained rectified flow model learns a direct flow from noise to clean images.
TReFT, by utilizing flow from noise to real images, is easier to finetune and converges faster.
The Inversion method adds an extra inversion step, increasing computation and slowing inference.
In practice, TReFT is the best option for its simplicity and speed.


\subsection{Applications On Pretrained RF Model}

\begin{figure}[t]
    \centering
    \includegraphics[width=1\linewidth]{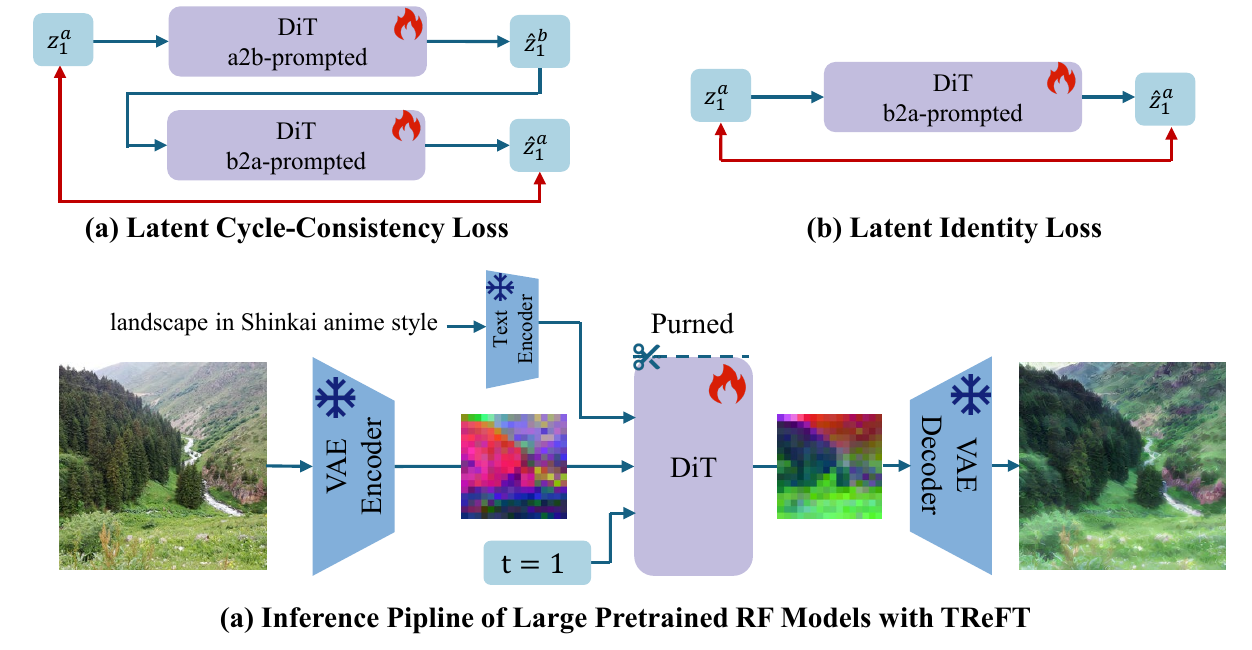}
    \caption{{\bf Application on RF models.} Top: Latent Cycle-Consistency Loss and Latent Identity Loss. Bottom: Inference pipline of Large Pretrained RF Models with TReFT.
    }
    \label{fig:figure-5}
\end{figure}

As Fig.~\ref{fig:figure-5} displays, we apply TReFT to pretrained RF models, freezing all modules except the MM-DiT block \cite{esser2024scaling}.
To preserve content on unpaired data while lowering memory use, we introduce latent cycle-consistency loss and identity loss.

\noindent{\bf Latent Cycle-Consistency Loss.}
Like in pixel space, we design a cycle-consistency loss in latent space.
The $z_1^a$ is fed into DiT with prompt a2b to get $\hat{z}_1^b$,
which is then passed back with prompt b2a to produce $\hat{z}_1^a$.
The original $z_1^a$ and reconstructed $\hat{z}_1^a$ form the latent cycle-consistency loss for a-to-b translation:
\begin{equation}
\begin{aligned}
    L_{cyc_a}&={E}_a \left[ \left\| DiT_{b \to a}(DiT_{a \to b}(z_1^a)) - z_1^a \right\|_1 \right] + \\
     &d_{LatentLPIPS}(DiT_{b \to a}(DiT_{a \to b}(z_1^a)),z_1^a).
\end{aligned}
\end{equation}
The latent cycle-consistency loss consists of an L1 loss and a LatentLPIPS loss \cite{kang2024distilling}.

\noindent{\bf Latent Identity Loss.}
To restrict DiT fine-tuning to the source domain, we design a latent identity loss.
Feeding $z_1^a$ with prompt b2a should output $\hat{z}_1^a$ close to $z_1^a$.
Thus, the latent identity loss for a-to-b translation is:
\begin{equation}
\begin{aligned}
    L_{idt_a}= &{E}_a \left[ \left\| DiT_{b \to a}(z_1^a) - z_1^a \right\|_1 \right] + \\
    &d_{LatentLPIPS}(DiT_{b \to a}(z_1^a),z_1^a).
\end{aligned}
\end{equation}

\noindent{\bf Adversarial Loss.}
We employ adversarial loss \cite{goodfellow2020generative} to supervise training.
The discriminator is designed based on Vision-Aided GAN: it utilizes a CLIP backbone to extract image features, followed by a simple MLP that classifies these features as real or fake.

\noindent{\bf Total Loss.}
The total loss can be represented as:
\begin{equation}
\begin{aligned}
    L= &\lambda_{cyc} \cdot L_{cyc} + \lambda_{idt} \cdot L_{idt} + \lambda_{gan} \cdot L_{gan},
\end{aligned}
\end{equation}
where $\lambda_{cyc}$, $\lambda_{idt}$ and $\lambda_{gan}$ are the weights of the three losses.

\noindent{\bf Lightweight modification.}
Visualization of VAE-decoded activations in SD3.5 reveals minimal changes in early MM-DiT blocks (Appendix Sec.~\ref{sec:sup_lightweight_modification}). Since early layers primarily extract visual features with little text influence, we convert them to standard DiT blocks by removing text branches to accelerate inference with minimal performance drop.
\section{Experiment}
\label{sec:experiment}


\subsection{Datasets}

\begin{figure*}[htb]
    \centering
    \includegraphics[width=0.75\linewidth]{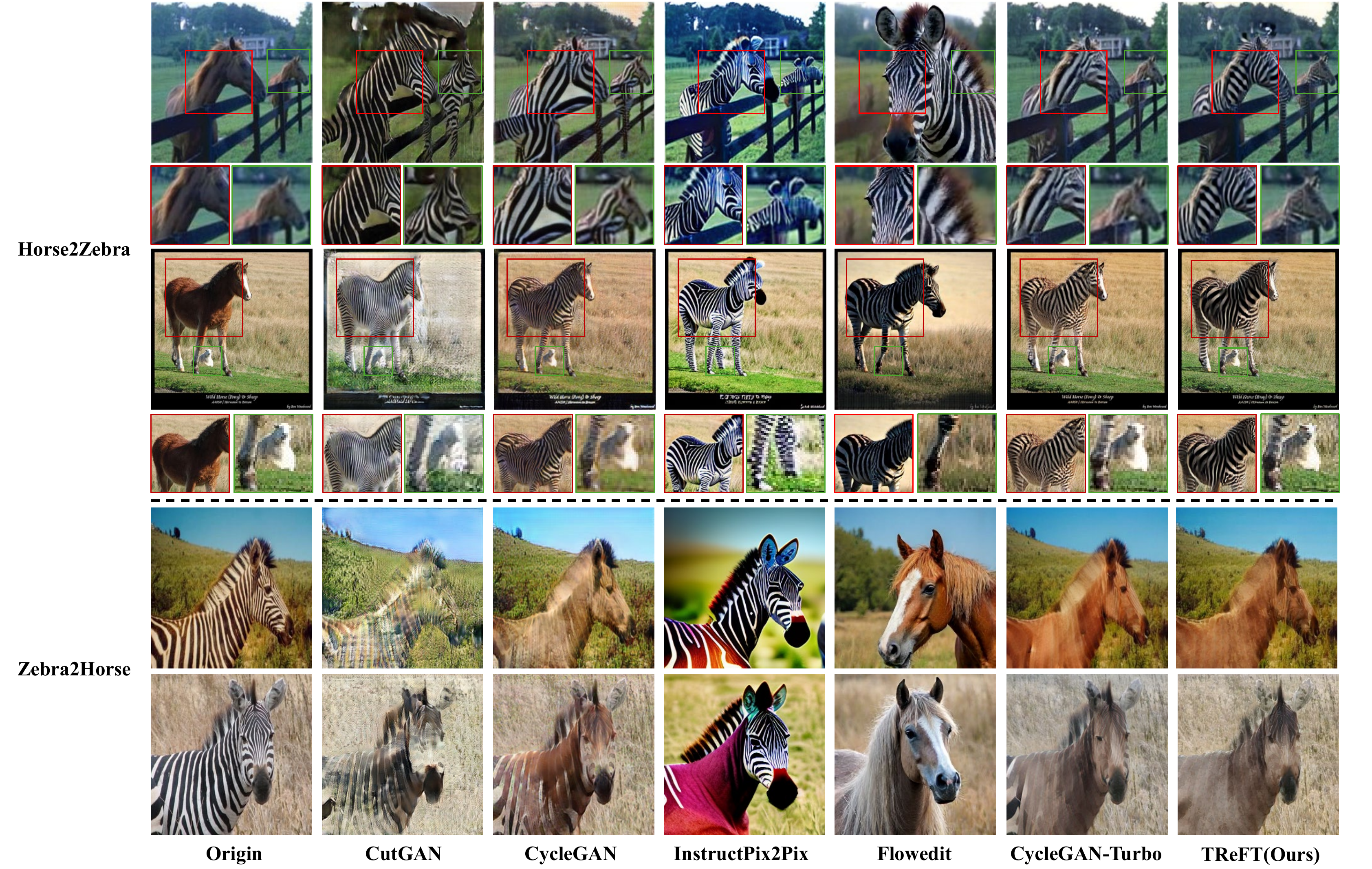}
    \caption{{\bf Comparison on Horse2Zebra dataset.} 
    The first column is origin image. 
    From left to the right are the results of CutGAN, CycleGAN, InstructPix2Pix, CycleGAN-Turbo, and TReFT (Ours).
    Zoom in to see details. More visual results are in Appendix Sec.~\ref{sec:sup_addtional_vis_results}.} 

    \label{fig:figure-6}
\end{figure*}

\begin{figure*}[htb]
    \centering
    \includegraphics[width=0.95\linewidth]{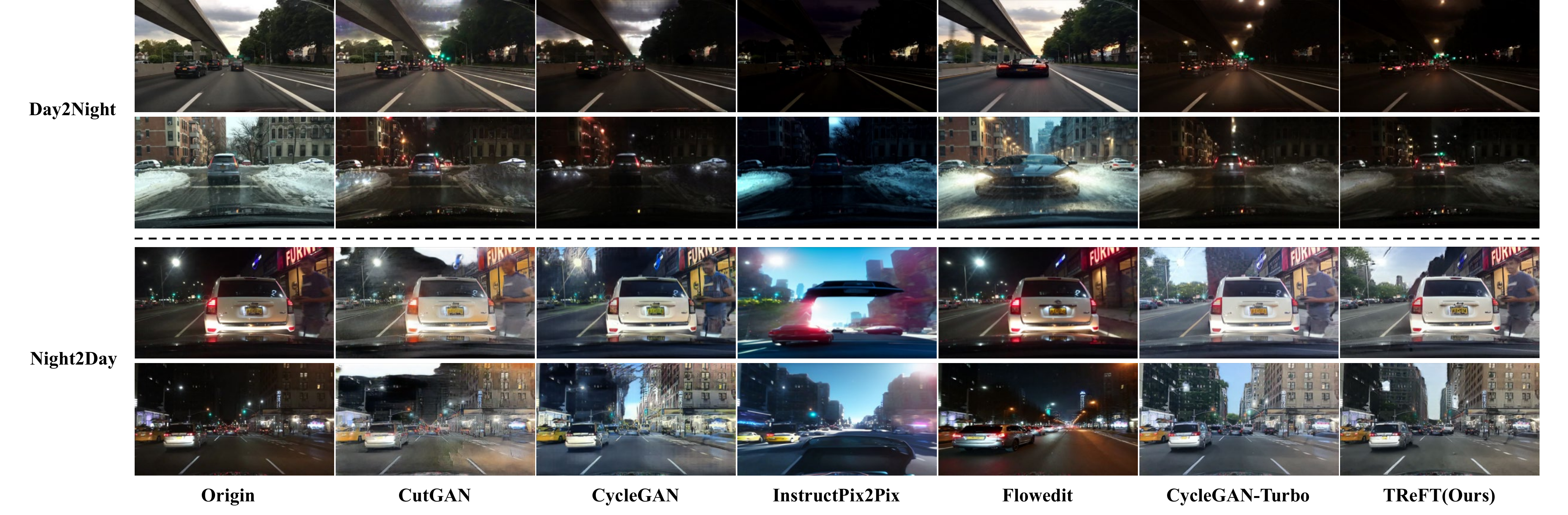}
    \caption{{\bf Comparison on BDD Day2Night dataset.} 
    The first column is origin image. Zoom in to see details.}

    \label{fig:figure-7}
\end{figure*}


We conduct extensive comparison and ablation experiments on both unpaired and paired datasets to validate the performance and effectiveness of TReFT.


\textbf{{Unpaired datasets.}} We mainly utilize the following unpaired datasets for our experiments:  
Horse2Zebra \cite{cyclegan}, BDD Day2Night \cite{yu2020bdd100k}, BDD Clear2Rainy \cite{yu2020bdd100k}, and LHQ2Shinkai \cite{jiang2023scenimefy}.

\textbf{{Paired datasets.}}  
We mainly utilize an artistic dataset collected from the community \cite{mid_journey_dataset}, and follow the preprocessing method of ControlNet \cite{controlnet} to obtain photo–edge pairs.

Detailed setting for datasets are provided in the Appendix Sec.~\ref{sec:sup_Experiment_Settings}.

\subsection{Implementation} 

We conduct all comparison experiments based on the pretrained RF models: SD3.5-Large-Turbo \cite{sauer2024fast} and FLUX.1-Schnell \cite{FLUX_website}.
We apply LoRA \cite{hu2022lora} to finetune only the linear layers in each MM-DiT block and the convolution layer in PatchEmbed module of DiT.

For the unpaired datasets, we set the batch size to 2 on two A800 gpus and the learning rate to 5e-6 with Adam optimizer.
The $\lambda_{cyc}$, $\lambda_{idt}$ and $\lambda_{gan}$ are set to 0.5, 1 and 1, respectively.
The lora rank of dit is chosen from 32, 64, 128, which is depend on the datasets.
To decrease the huge consumption of GPU memory, we use the mix-precision of bf16 for dit and fp32 for other modules.

For the paired dataset edge2photo, we set the batch size to 8 on eight A800 gpus and the learning rate to 1e-5 with Adam optimizer.
Following the setting of pix2pix, we removed the constraint of latent losses, and use
 the clip score and the lpips net to constraint the paired image in pixel level.
The lora rank is set to 512 and the gan loss is set to 0.5.
We also use bf16 for DiT to reduce the memory consumption.

\subsection{Evaluation Metrics}

For unpaired datasets, we use FID \cite{heusel2017gans} to evaluate translation quality and the DINO-Struct score \cite{tumanyan2022splicing, img2img-turbo} to assess content preservation. Note that the DINO Struct scores are multiplied by 100 in our paper.

\subsection{Comparison Experiment}





\noindent{\bf Baselines.}
For the unpaired datasets, we select multiple baselines, ranging from GAN-based models to diffusion-based models.
All the baselines can be grouped into two categories: GAN-based models, diffusion-based models.
For the paired dataset, we compare SD3.5 with TReFT trained on edge2photo with SD3.5-ControlNet.

\noindent{\bf Quantitative analysis.}
As shown in Table~\ref{tab:tab-1}, TReFT (on SD3.5-large-turbo and FLUX.1-Schnell) achieves high-quality image generation with low FID and DINO Struct scores. It attains SOTA FID on Day2Night and Night2Day, and second-best DINO Struct on both. Plug\&Play and InstructPix2Pix achieve the best DINO Struct but with much worse FID, showing poor domain translation. On Horse2Zebra, our method achieves the best performance considering both FID and DINO Struct.

\noindent{\bf Qualitative analysis.}
As shown in Fig.~\ref{fig:figure-6} and Fig.~\ref{fig:figure-7}, TReFT is capable of generating images that closely resemble those from the target domain while effectively preserving the content of the original images.
For example, on the Day2Night dataset, our model generates scenes with fewer lighting artifacts and avoids duplicated moons, issues that are evident in other models.
On the Night2Day task, our model successfully avoids the common artifact of mixing buildings with trees, a problem frequently seen in the outputs of CycleGAN-Turbo and other approaches.

\begin{figure}[t]
    \centering
    \includegraphics[width=0.95\linewidth]{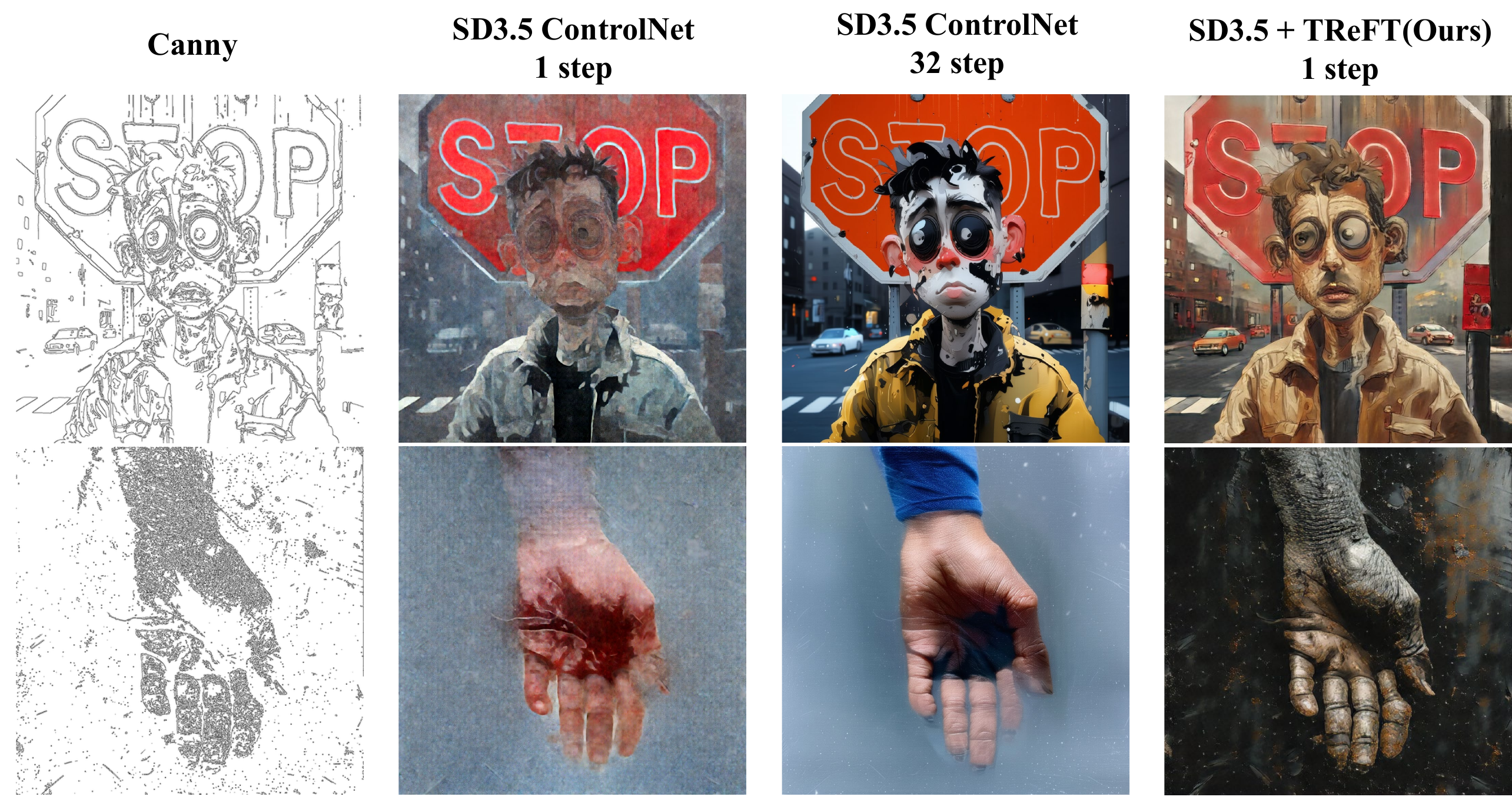}
    \caption{{\bf Compared with SD3.5 ControlNet on the Edge2Photo task.}
    From left to right are respectively Canny edge input, SD3.5 ControlNet 1-step, SD3.5 ControlNet 32-step, TReFT(Ours) 1-step.
    Zoom in to see details.}

    \label{fig:figure-8}
\end{figure}

Additionally, we compare the performance of our method with SD3.5-ControlNet on the paired dataset edge2photo.
As Fig.~\ref{fig:figure-8} shows,
our method can generate high-quality images in a single step, in contrast to SD3.5-ControlNet, which requires 32 steps to achieve comparable results.

\subsection{Ablation Study}

\vspace{0.3em}
\noindent{\bf Ablation of Vanilla, Inversion, and TReFT.}
To further validate the effectiveness of TReFT,
we conduct ablation studies using two pretrained RF models on two unpaired datasets: Horse2Zebra and Day2Night.
All three methods are trained for 16k steps, and we report the best FID and DINO-Struct scores achieved.
As shown in Table~\ref{tab:tab-2}, Inversion and TReFT exhibit similar performance on both metrics across the two datasets.
Meanwhile, the Vanilla method performs poorly in terms of FID and also achieves relatively lower DINO-Struct scores compared to the other two.
This indicates that the Vanilla method tends to preserve the input image with minimal modifications in most cases, thereby failing to effectively accomplish the image translation task.

\begin{table}[t]
  \centering
  \footnotesize
  {
    
  \begin{tabular}{c|cccc}
    \toprule
    \multirow{3}{*}{Method}                                                      & \multicolumn{2}{l}{Horse $\rightarrow$ Zebra}                                       & \multicolumn{2}{l}{Day $\rightarrow$ Night}                                         \\
                                                                                 & FID$\downarrow$ & \begin{tabular}[c]{@{}l@{}}Dino\\ Struct$\downarrow$\end{tabular} & FID$\downarrow$ & \begin{tabular}[c]{@{}l@{}}Dino\\ Struct$\downarrow$\end{tabular} \\
    \midrule
    
    SD3.5+Vanilla                                                                & 117.3           & 1.1                                                               & 57.3            & 4.8                                                               \\
    \cmidrule{2-5}
    SD3.5+Inversion                                                              & 41.3            & 2.4                                                               & 30.6            & 2.6                                                               \\
    SD3.5+TReFT (Ours)                                                            & 39.2            & 2.4                                                               & 30.9            & 2.9                                                               \\
    \midrule[0.2pt]
    FLUX.1+Vanilla                                                               & 97.3            & 1.7                                                               & 49.6            & 3.1                                                               \\
    \cmidrule{2-5}
    FLUX.1+Inversion                                                             & 40.3            & 3.1                                                               & 31.7            & 3.3                                                               \\
    FLUX.1+TReFT (Ours)                                                           & 46.6            & 2.7                                                               & 29.9            & 2.8                                                               \\
    
    \bottomrule
  \end{tabular}
  }
  \caption{{\bf Ablation on different training method.}}
  \label{tab:tab-2}
\end{table}


 \begin{table}[t]
  \centering
  \small
  {
    
  \begin{tabular}{ccc}
    \toprule
    \multirow{2}{*}{Method}         & \multicolumn{2}{c}{Horse $\rightarrow$ Zebra} \\
                                    & FID$\downarrow$   & DINO Struct$\downarrow$   \\
    \midrule
    Without Constraint& 37.9         & 7.6                       \\
    \midrule[0.2pt]
    Pixel space  & 40.9 (+7.9\%)              & 3.7 (-51.3\%)                       \\
    Latent space & 38.3 (+1.1\%)              & 2.7 (-64.5\%)                       \\
    \bottomrule  
  \end{tabular}
  }
  \caption{{\bf Ablation study of losses in latent space level.}The percentage is computed with respect to the metrics of the unconstrained method.}
  \label{tab:tab-3}
  \vspace{-0.5em}
\end{table}

\noindent{\bf Ablation study of losses in latent space.}
To examine the effect of losses at the latent space level, we evaluate model performance using cycle-consistency and identity losses applied in both the latent space and the pixel space \cite{cyclegan}.
For comparison, we use a baseline model without these two loss constraints.
As shown in Table~\ref{tab:tab-3}, losses applied in the latent space are more effective than those in the pixel space, achieving a greater reduction in DINO-Struct with a smaller increase in FID.
We attribute this to the fact that latent-space losses can directly optimize the parameters of DiT without passing through the VAE decoder.

\begin{table}[t]
  \centering
  \footnotesize
    {
    
  \begin{tabular}{cccc}
    \toprule
    \multirow{2}{*}{Single Block Num}    & \multirow{2}{*}{\begin{tabular}[c]{@{}l@{}}Inference\\ Time (ms)$\downarrow$\end{tabular}}     & \multicolumn{2}{c}{Horse $\rightarrow$ Zebra} \\
                            &    & FID$\downarrow$   & Dino Struct$\downarrow$  \\
    \midrule
    0                     & 157   & 38.3           & 2.7                             \\
    12                    & 139   & 38.5           & 2.2                    \\
    18                    & 132   & 38.1           & 2.9                  \\
    24                    & 124   & 40.8           & 2.8                  \\
    30                    & 116   & 42.2           & 2.8                  \\
    36                    & 108   & 44.3           & 2.7                  \\
    38*                   & 101   & 46.1           & 2.9                   \\
    \midrule
    CycleGAN-Turbo\cite{img2img-turbo} & 135   & 41.0           & 2.1                   \\
    \bottomrule  
  \end{tabular}
  }
  \caption{{\bf Ablation study of lightweight modification for MM-DiT on SD3.5.} SD3.5 has 38 MM-DiT blocks in total.}
  \label{tab:tab-4}
  \vspace{-0.5em}
\end{table}

\vspace{0.3em}
\noindent{\bf Ablation study of lightweight modification for MM-DiT.}
To assess the impact of our lightweight design on model performance, we conduct an ablation study on SD3.5-Large-Turbo by varying the number of early MM-DiT blocks replaced with single blocks that omit the text branch. As shown in Table~\ref{tab:tab-4}, replacing up to the first 18 MM-DiT blocks introduces minimal performance degradation while providing notable inference speedup. Inference time is measured on an A800 GPU.

\section{Conclusion}
\label{sec:conclusion}


In this work, we investigated the convergence challenges of fine-tuning RF models for one-step image translation and identified the objective mismatch between RF and diffusion models as the main cause of instability. We further proved that the predicted velocity converges to the clean image near the end of denoising. Building on these insights, we proposed {\bf TReFT}, a simple yet effective strategy that stabilizes adversarial fine-tuning with one-step inference. With additional engineering optimizations, TReFT enables real-time inference and achieves comparable performance with sota methods on multiple benchmarks.

{
    \small
    \bibliographystyle{ieeenat_fullname}
    \bibliography{main}
}

\clearpage
\setcounter{page}{1}
\maketitlesupplementary

\section{Experiment details for Fig.~\ref{fig:figure-1}}
\label{sec:sup_exp_fig_1}

To examine the cause of convergence issues, we conduct ablation experiment on Horse2zebra dataset.
The training curves is displayed in Fig.~\ref{fig:figure-1} in the main paper.
Specifically, we compared SD-Turbo\cite{sauer2024adversarial} and PixArt-Alpha\cite{chen2023pixart} (which differ in backbone), as well as SD2.1\cite{rombach2022high} and its PeRFlow-finetuned variant \cite{yan2024perflow} (which differ in training objective), using the Vanilla fine-tuning method as in CycleGAN-Turbo.

For all models, We set the batch size to 2 on two A800 gpus and the learning rate to 1e-5 with Adam optimizer.
The $\lambda_{cyc}$, $\lambda_{idt}$ and $\lambda_{gan}$ are set to 1, 1 and 1, respectively.
For FLUX.1-Schnell and PixArt-Alpha, the lora rank of DiT is 128. For SD-Turbo, SD2.1 and its PeRFlow-finetuned variant, the lora rank of UNet is 128.


\section{Datasets Settings Details}
\label{sec:sup_Experiment_Settings}

\noindent\textit{Unpaired datasets.} We mainly ultilize the following paired datasets for experiments:
\begin{itemize}
\item Horse2zebra\cite{cyclegan}. Following CycleGAN, we use the 939 images form wild
horse class and 1,177 images from the zebra class in Imagenet. 
For this dataset, use load the 286$\times$286 images and do 256$\times$256 center crops when training.
During inference, we directly apply translation at 256$\times$256.
All the metrics is calculated on the full validation set of Horse2zebra.

\item BDD Day2Night\cite{yu2020bdd100k}. We use the Day and Night subsets of the BDD100k dataset.
Following CycleGAN, we resize all the images to 512$\times$512 during the training and inference.
The metrics is calculated on the validation set of it.

\item BDD Clear2Rainy\cite{yu2020bdd100k}. We use the Clear and Rainy subsets of the BDD100k dataset.
Its setting is the same as BDD Day2Night.

\item LHQ2Shinkai\cite{jiang2023scenimefy}. This dataset is a filtered version of LHQ and Shinkai.
To enhance aesthetics, we filtered 2,000 images from Landscapes High-Quality (LHQ) dataset and 1,748
images form the Shinkai dataset.

\end{itemize}

\noindent\textit{Paired datasets.}
 We mainly ultilize a artistic dataset collect from the community \cite{mid_journey_dataset}, and follow the 
 pre-process of ControlNet to get photo and edge paires.

\section{Proof for Theorem 1}
\label{sec:sup_proof_theorem_1}

In the step-by-step denoising process of a text-conditioned RF model, let

\begin{equation}
z_1 \sim \mathcal N(\mu,\sigma^2 I_d),\qquad z_0 \sim \mathcal N(0, I_d),
\end{equation}

where $z_0$ and $z_1$ are independent.  
Given the intermediate latent

\begin{equation}
z_t = (1-t) z_0 + t z_1,\qquad t\in(0,1),
\end{equation}

our goal is to compute the conditional expectation $E[z_1 - z_0 \mid z_t]$.

Define the joint Gaussian vector

\begin{equation}
Z = \begin{bmatrix} z_1 \\ z_t \end{bmatrix}.
\end{equation}

Since $z_t$ is a linear combination of independent Gaussian variables, the pair $(z_1, z_t)$ is jointly Gaussian. Hence the conditional expectation $E[z_1 \mid z_t]$ admits the standard linear-Gaussian form.  
We first compute the mean of $z_t$ and the covariance terms involving $z_1$ and $z_t$:



\begin{equation}
E[z_t] = t \mu + (1 - t) \cdot 0 = t \mu, \quad E[Z] = \begin{bmatrix} \mu \\ t \mu \end{bmatrix},
\end{equation}

\begin{equation}
\text{Cov}(z_1, z_t) = t \cdot \text{Cov}(z_1, z_1) = t \sigma^2 I,
\end{equation}

\begin{equation}
\text{Cov}(z_t, z_t) = t^2 \sigma^2 I + (1 - t)^2 I.
\end{equation}

Thus, the joint distribution of $Z$ is:

\begin{equation}
Z = \begin{bmatrix} z_1 \\ z_t \end{bmatrix} \sim \mathcal{N}\left( \begin{bmatrix} \mu \\ t\mu \end{bmatrix}, \begin{bmatrix} \sigma^2 I & t \sigma^2 I \\ t \sigma^2 I & t^2 \sigma^2 I \!+\! (1 \!-\! t)^2 I \end{bmatrix} \right).
\end{equation}

Using the standard formula for the conditional expectation of a multivariate Gaussian:

\begin{equation}
E[z_1 \mid z_t] = \mu + t \sigma^2 \left( t^2 \sigma^2 + (1 - t)^2 \right)^{-1} (z_t - t \mu).
\end{equation}

Similarly, we can derive $E[z_0 \mid z_t]$ by constructing the joint distribution of $[z_0, z_t]^\top$:

\begin{equation}
E[z_0 \mid z_t] = (1 - t) \left( t^2 \sigma^2 + (1 - t)^2 \right)^{-1} (z_t - t\mu).
\end{equation}

By linearity of conditional expectation:

\begin{equation}
E[z_1 - z_0 \mid z_t] = E[z_1 \mid z_t] - E[z_0 \mid z_t].
\end{equation}

Substituting the expressions derived above and simplifying:

\begin{equation}
E[z_1 \!-\! z_0 \mid z_t] = \frac{t\sigma^2 \!-\! (1 \!-\! t) }{t^2\sigma^2 \!+\! (1 \!-\! t)^2 } z_t + \frac{(1 \!-\! t) }{t^2\sigma^2 \!+\! (1 \!-\! t)^2 } \mu.
\label{eq:sp-eq-8}
\end{equation}

\section{Proof for Theorem 2}
\label{sec:sup_proof_theorem_2}

We restate a precise version of Theorem~2 and then give a rigorous derivation based on a local Laplace expansion under a $C^{1,1}$ condition.

\bigskip  
\noindent\textbf{Theorem 2.}  
Let $z_0\sim\mathcal N(0,I_d)$ and let $z_1\sim P(z_1\mid c)$. Fix a realized clean latent $z_1^*$ drawn from $P(z_1\mid c)$. Denote $\tau := 1-t\in(0,1)$ and assume the observed intermediate latent is given by

\begin{equation}
z_t = t z_1^* + \tau z_0.
\end{equation}

Suppose there exists an open neighborhood $U$ of $z_1^*$ and constants $t_0\in(0,1)$, $L>0$ such that for all $t\in[t_0,1)$ the following hold:

1.  (Positivity) $p(z_1\mid c)>0$ for all $z_1\in U$.
    
2.  (Local $C^{1,1}$) $\log p(\cdot\mid c)$ is continuously differentiable on $U$ and its gradient is Lipschitz with constant $L$: for all $x,y\in U$,

\begin{equation}
\|\nabla\log p(x\mid c) - \nabla\log p(y\mid c)\| \le L \|x-y\|.
\end{equation}

3.  (MLE in $U$) For sufficiently small $\tau$ the MLE $\hat z := z_t/t$ lies in $U$, and the local quadratic term induced by the likelihood is non-degenerate (equivalently $t$ is bounded below by $t_0>0$).

Then, the posterior mean $m(z_t):=\mathbb E[z_1\mid z_t]$ satisfies: 

\begin{equation}
m(z_t) \;=\; \hat z + O(\tau^2),
\end{equation}

and consequently

\begin{equation}
\mathbb E[z_1 - z_0\mid z_t] \;=\; z_1^* + O(\tau),
\end{equation}

as $\tau\to0$ (equivalently $t\to1$). In particular $\lim_{t\to1}\mathbb E[z_1 - z_0\mid z_t] = z_1^*$.

\bigskip  
\noindent\textbf{Proof.}

\paragraph{S1: Change of variables and posterior expression.}  
Write $g(z):=\log p(z\mid c)$. Given the observation $z_t$, the posterior density of $z_1$ is

\begin{equation}
p(z_1\mid z_t) \propto \exp\big( g(z_1) \big) \cdot \exp\!\Big(-\frac{1}{2\tau^2}\|z_t - t z_1\|^2\Big).
\end{equation}

Set $\hat z := z_t/t$. 
Change variables:

\begin{equation}
z_1 = \hat z + \tau u,\qquad u\in\mathbb R^d.
\end{equation}

Under this transform the likelihood factor becomes

\begin{equation}
\exp\Big(-\frac{1}{2\tau^2}\|z_t - t z_1\|^2\Big) = \exp\Big(-\frac{t^2}{2}\|u\|^2\Big).
\end{equation}

Thus the posterior density of $u$ (up to normalization) is

\begin{equation}
p_u(u) \propto \exp\Big( g(\hat z + \tau u) - \tfrac{t^2}{2}\|u\|^2 \Big).
\end{equation}

\paragraph{S2: First-order Taylor and gradient-Lipschitz remainder bound.}  
Apply a first-order Taylor expansion of $g$ at $\hat z$ and bound the remainder using the gradient-Lipschitz property. For any $u$ with $\hat z+\tau u\in U$,

\begin{equation}
g(\hat z + \tau u) = g(\hat z) + \tau \nabla g(\hat z)^\top u + R(\tau,u),
\end{equation}

with the remainder controlled by

\begin{equation}
|R(\tau,u)| \le \frac{L}{2}\tau^2 \|u\|^2.
\end{equation}

\begin{mdframed}
\noindent\textbf{Lemma 2.1 (L-smooth remainder bound).}
If $g\in C^{1,1}(U)$ with gradient Lipschitz constant $L$, then
\[
| g(x+h) - g(x) - \nabla g(x)^\top h |
\le \frac{L}{2}\|h\|^2.
\]
(Proof: by the mean value theorem and Lipschitz gradient.)
\end{mdframed}

Hence

\begin{equation}
p_u(u) \propto \!\exp\Big( g(\hat z) \!+\!\tau \nabla g(\hat z)^\top u \!-\! \tfrac{t^2}{2}\|u\|^2 \!+\! R(\tau,u) \Big).
\end{equation}

Dropping the constant $g(\hat z)$ (absorbed into normalization), we may write

\begin{equation}
p_u(u) \propto \exp\Big( -\tfrac{1}{2} u^\top A u + b^\top u + r(u)\Big),
\end{equation}

where

\begin{equation}
A := t^2 I_d,\quad b := \tau \nabla g(\hat z),\quad |r(u)| \le \tfrac{L}{2}\tau^2\|u\|^2.
\end{equation}

\paragraph{S3: Dominant Gaussian and perturbative expansion for the mean.}  
If $r(u)\equiv0$ (no remainder), then $p_u$ is exactly Gaussian with precision $A=t^2 I$ and linear term $b$, so its mean would be

\begin{equation}
\mathbb E_G[u] = A^{-1} b = t^{-2} (\tau \nabla g(\hat z)).
\end{equation}

In the presence of the small remainder $r(u)$ satisfying $|r(u)| \le \tfrac{L}{2}\tau^2\|u\|^2$, one can view the true density as the above Gaussian density multiplied by a factor $\exp(r(u))$ that is uniformly close to $1$ for $\tau$ small on any region where $\|u\|$ is $O(1)$. We denote that:

\begin{equation}
\begin{aligned}
    m_u=&\mathbb E_{\mathrm G}[u]=t^{-2}(\tau\nabla g(\hat z)), \\
    \Sigma_u=&\operatorname{Cov}_{\mathrm G}(u)=t^{-2}I_d, \\
    r(u)=&\tau^2 s(u), \quad with \, |s(u)|\le~\tfrac{L}{2}\|u\|^2.
\end{aligned}
\end{equation}







We start from

\begin{equation}
\mathbb E[u] \;=\; \frac{\displaystyle\int u\,p_{\mathrm G}(u)\,e^{r(u)}\,du} {\displaystyle\int p_{\mathrm G}(u)\,e^{r(u)}\,du} =\frac{N}{D}.
\end{equation}

Expand $e^{r(u)}$ to second order in $r$:

\begin{equation}
e^{r(u)} = 1 + r(u) + \tfrac12 r(u)^2 + O(r(u)^3).
\end{equation}

Since $r(u)=\tau^2 s(u)$ with $|s(u)|\le \tfrac{L}{2}\|u\|^2$, we have $r(u)=O(\tau^2\|u\|^2)$ and $r(u)^2=O(\tau^4\|u\|^4)$.

\medskip  
\noindent\textbf{For the numerator.}

\begin{equation}
\begin{aligned} N =& \int u\,p_{\mathrm G}(u)\big(1 + r(u) + \tfrac12 r(u)^2 + \cdots\big)\,du \\ 
=& \underbrace{\int u\,p_{\mathrm G}(u)\,du}_{=m_u} \;+\; \int u\,p_{\mathrm G}(u) r(u)\,du \\
&\;+\; O\!\big(\mathbb E_{\mathrm G}[\|u\|\,r(u)^2]\big). 
\end{aligned}
\end{equation}

Estimate the second term:

\begin{equation}
\int u\,p_{\mathrm G}(u) r(u)\,du = \tau^2 \int u\,p_{\mathrm G}(u) s(u)\,du.
\end{equation}

Using $|s(u)|\le \tfrac{L}{2}\|u\|^2$ and the Gaussian moment identity (for $u\sim\mathcal N(m_u,\Sigma_u)$)

\begin{equation}
\mathbb E_{\mathrm G}[\,u\|u\|^2\,] = m_u\big(\|m_u\|^2 + \operatorname{tr}\Sigma_u\big) + 2\Sigma_u m_u,
\end{equation}

we see $\mathbb E_{\mathrm G}[u\|u\|^2]=O(m_u)+O(\Sigma_u m_u)=O(\tau)$ because $m_u=O(\tau)$ and $\Sigma_u=O(1)$. Hence

\begin{equation}
\begin{aligned}
    \int u\,p_{\mathrm G}(u) r(u)\,du = \tau^2\cdot O(\tau)=O(\tau^3).
\end{aligned}
\end{equation}

The remainder term $\mathbb E_{\mathrm G}[\|u\|\,r(u)^2]=O(\tau^4)$ is higher order. Therefore

\begin{equation}
N = m_u + O(\tau^3).
\end{equation}

\medskip
\noindent\textbf{For the denominator.}

\begin{equation}
\begin{aligned} D &= \int p_{\mathrm G}(u)\big(1 + r(u) + \tfrac12 r(u)^2 + \cdots\big)\,du \\ &= 1 + \int p_{\mathrm G}(u) r(u)\,du + O(\mathbb E_{\mathrm G}[r(u)^2]). \end{aligned}
\end{equation}

Here

\begin{equation}
\int p_{\mathrm G}(u) r(u)\,du = \tau^2 \int p_{\mathrm G}(u) s(u)\,du = \tau^2\cdot O(1),
\end{equation}

since $\int p_{\mathrm G}\|u\|^2 = \|m_u\|^2 + \operatorname{tr}\Sigma_u = O(1)$. 

Also $\mathbb E_{\mathrm G}[r(u)^2]=O(\tau^4)$. Thus

\begin{equation}
\begin{aligned}
    D =& 1 + c\,\tau^2 + O(\tau^4), \\
    where \quad c =& \int p_{\mathrm G}(u) s(u)\,du = O(1).
\end{aligned}
\end{equation}

\smallskip
\noindent\textbf{For the ratio.}

Use the expansion:

\begin{equation}
\begin{aligned}
    \frac{m_u \!+\! O(\tau^3)}{1 \!+\! c\tau^2 \!+\! O(\tau^4)} =& (m_u \!+\! O(\tau^3))\big(1 \!- \!c\tau^2 \!+\! O(\tau^4)\big) \\
    =& m_u + m_u\cdot O(\tau^2) + O(\tau^3).
\end{aligned}
\end{equation}

Since $m_u=O(\tau)$, we have $m_u\cdot O(\tau^2)=O(\tau^3)$. Hence

\begin{equation}
\mathbb E[u] = m_u + O(\tau^3) = t^{-2}(\tau\nabla g(\hat z)) + O(\tau^3).
\end{equation}

\paragraph{S4: Posterior mean of $z_1$.}  
Returning to $z_1=\hat z+\tau u$, we obtain

\begin{equation}
\begin{aligned}
m(z_t) :=& \mathbb E[z_1\mid z_t] = \hat z + \tau \mathbb E[u] \\
=& \hat z + \tau\big( t^{-2}\tau \nabla g(\hat z) + O(\tau^3)\big) \\
=& \hat z + O(\tau^2).
\end{aligned}
\end{equation}

\paragraph{S5: Relation to the desired quantity $\mathbb E[z_1-z_0\mid z_t]$.}  
By algebra from the generative model,

\begin{equation}
z_t = t z_1 + \tau z_0 \quad\Rightarrow\quad z_1 - z_0 = \frac{z_1 - z_t}{\tau}.
\end{equation}

Taking conditional expectation given $z_t$ yields the identity

\begin{equation}
\mathbb E[z_1 - z_0\mid z_t] \;=\; \frac{m(z_t) - z_t}{\tau}.
\end{equation}

Substitute $m(z_t)=\hat z + O(\tau^2)$ and $\hat z = z_t / t$:

\begin{equation}
\begin{aligned}
    \mathbb E[z_1 - z_0\mid z_t] =& \frac{\hat z - z_t}{\tau} + O(\tau) \\
    =& \frac{z_t/t - z_t}{\tau} + O(\tau) \\
    =& \frac{z_t}{t} + O(\tau).
\end{aligned}
\end{equation}

Finally, since $z_t/t = z_1^* + (\tau/t) z_0$ and $(\tau/t) z_0 = O(\tau)$, we obtain

\begin{equation}
\mathbb E[z_1 - z_0\mid z_t] = z_1^* + O(\tau),
\end{equation}

Taking $\tau\to0$ gives $\lim_{t\to1}\mathbb E[z_1 - z_0\mid z_t] = z_1^*$.

\begin{figure*}[htb]
    \centering
    \includegraphics[width=0.95\linewidth]{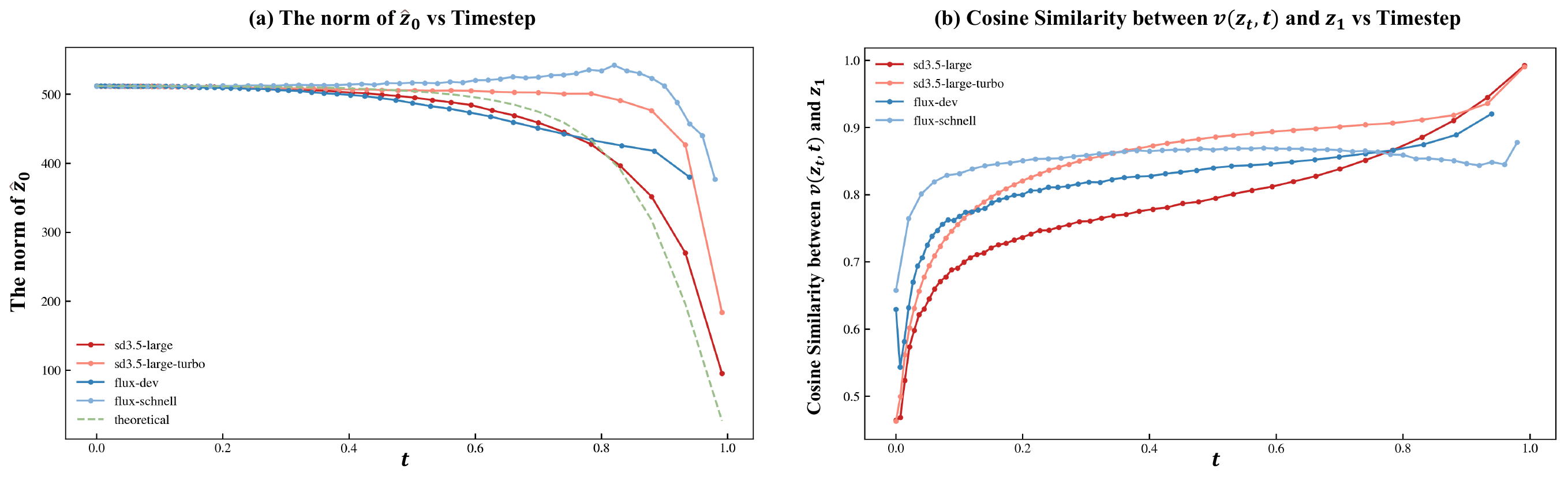}
    \caption{"The norm of $\hat{z}_0$ vs Timestep" and "Cosine Similarity between $z_1$ and $v_{\theta}$ vs Timestep".
   The curves in (a) are SD3.5-Large, SD3.5-Large-Turbo, FLUX.1-Schnell, FLUX.1-Dev, and theory Curve for "The norm of $\hat{z}_0$ vs Timestep".}
   \label{fig:sup_fig_10}
\end{figure*}

\section{Experiment On Theorem 1 and 2}
\label{sec:sup_exp_theorem}

In Observation 2 of Sec.~\ref{sec:treft}, we observed that as the timestep increases,
the noise level in the pretrained DiT's direct output gradually decreases.
This is generally true for all the pretrained DiT models.
We conduct an experiment to verify this phenomenon on SD3.5-Large, SD3.5-Large-Turbo, FLUX.1-Dev and FLUX.1-Schnell.
Specifically, we generate 1000 prompts by LLM and simulate the generation process
 of the each pretrained DiT model for 50 inference steps to generate $1024 \times 1024$ images.
We measure cosine similarity between the predicted velocity at each timestep and the final image to evaluate their difference, and compute the norm of the noise vector to analyze noise components, which is predicted via one-step inversion:
\begin{equation}
    \hat{z}_0=z_t - t v_{\theta}\left(z_t, t\right).
\label{eq:sp-eq-9}
\end{equation}
Fig.~\ref{fig:sup_fig_10} shows the results of our experiment.

\subsection{Deriving the Expected Norm of \texorpdfstring{$\hat{z_0}$}{z0}}

Under the same setting as Theorem~1, we aim to derive the norm of $\hat{z}_0$.
From previous derivations (see also Theorem 1), we have Eq.~\ref{eq:sp-eq-8}.
Let us denote:

\begin{equation}
    \lambda = \frac{1}{t^2\sigma^2 + (1 - t)^2 }.
\label{eq:sp-eq-10}
\end{equation}

Under L2 loss, the nueral network approximate the expectation of its target, that is:

\begin{equation}
    v_{\theta}\left(z_t, t\right) = E[z_1 \!-\! z_0 \mid z_t].
\label{eq:sp-eq-11}
\end{equation}

Combine equations Eq.~\ref{eq:sp-eq-8}, Eq.~\ref{eq:sp-eq-9}, Eq.~\ref{eq:sp-eq-10} and Eq.~\ref{eq:sp-eq-11}, we can derive that:

\begin{equation}
    \hat{z}_0 = (1-\lambda t^2 \sigma^2 + \lambda t (1-t))z_t + \lambda t (t-1)\mu.
\end{equation}

This can be rewritten as:

\begin{equation}
    \begin{aligned}
    &\hat{z}_0 = \alpha z_t + \beta\mu, \\
    where \quad &\alpha = (1-\lambda t^2 \sigma^2 + \lambda t (1-t)), \\
    and \quad &\beta = \lambda t (t-1).
\end{aligned}
\end{equation}

Then, the expected squared norm can be expanded as:

\begin{equation}
    E \left[ \left\| \hat{z_0} \right\|^2 \right] = \alpha^2 E[\|z_t\|^2] + 2 \alpha \beta E[z_t^\top \mu] + \beta^2 \|\mu\|^2.
\end{equation}

Expectation of $\|z_t\|^2$:
    
\begin{equation}
    \begin{aligned}
        E[\|z_t\|^2] =& E[z_t^\top z_t] \\
        =& E[(t z_1 \!+\! (1 \!-\! t) z_0)^\top (t z_1 \!+\! (1 \!-\! t) z_0)] \\
=& t^2 E[z_1^\top z_1] + 2 t (1 - t) E[z_1^\top z_0] + \\
&(1 - t)^2 E[X_0^\top X_0].
    \end{aligned}
\end{equation}

Since $z_1$ and $z_0$ are independent and $E[z_0] = 0$, the cross term vanishes:

\begin{equation}
    E[z_1^\top z_0] = {E}[z_1]^\top {E}[z_0] = \mu^\top 0 = 0.
\end{equation}

Thus,

\begin{equation}
    E[\|z_t\|^2] = t^2 {E}[z_1^\top z_1] + (1 - t)^2 {E}[z_0^\top z_0].
\end{equation}

For $z_1 \sim \mathcal{N}(\mu, \sigma^2 I_d)$, it is known that

\begin{equation}
\begin{aligned}
        {E}[z_1^\top z_1] &= \mathrm{tr}(\mathrm{Cov}(z_1)) + \|{E}[z_1]\|^2 \\
        &= d \sigma^2 + \|\mu\|^2.
\end{aligned}
\end{equation}

For $z_0 \sim \mathcal{N}(0, I_d)$,

\begin{equation}
    {E}[z_0^\top z_0] = \mathrm{tr}(I_d) + \|0\|^2 = d.
\end{equation}

Hence,

\begin{equation}
 {E}[\|z_t\|^2] = t^2 (\|\mu\|^2 + d \sigma^2) + (1 - t)^2 d. 
\end{equation}

Expectation of the inner product $E[z_t^\top \mu]$:
    
\begin{equation}
\begin{aligned}
{E}[z_t^\top \mu] =& {E}[(t z_1 + (1 - t) z_0)^\top \mu] \\
=& t {E}[z_1^\top \mu] + (1 - t) {E}[z_0^\top \mu].
\end{aligned}
\end{equation}

Since

\begin{equation}
{E}[z_1^\top \mu] = {E}[z_1]^\top \mu = \mu^\top \mu = \|\mu\|^2,
\end{equation}

and

\begin{equation}
{E}[z_0^\top \mu] = {E}[z_0]^\top \mu = 0^\top \mu = 0,
\end{equation}

we have

\begin{equation}
 {E}[z_t^\top \mu] = t \|\mu\|^2. 
\end{equation}

The final form for the expected norm of $\hat{z_0}$ is:

\begin{equation}
\begin{aligned}
E \left[ \left\| \hat{z_0} \right\|^2 \right] =& \alpha^2 {E}[\|z_t\|^2] + 2\alpha\beta {E}[z_t^\top \mu] + \beta^2 \|\mu\|^2 \\
where \quad \lambda =& \frac{1}{t^2\sigma^2+(1-t)^2} \\
\alpha =& 1 - \lambda t^2 \sigma^2 +\lambda t (1-t) \\
\beta =& \lambda t (t - 1) \\
{E}[\|z_t\|^2] =& t^2 (\|\mu\|^2 + d\sigma^2) + (1 - t)^2 d \\
{E}[z_t^\top \mu] =& t \|\mu\|^2
\end{aligned}
\end{equation}

\subsection{Discussion}



The empirical results are shown in Fig.~\ref{fig:sup_fig_10}. As illustrated in Fig.~\ref{fig:sup_fig_10}(a), for SD3.5-Large, FLUX.1-Dev, and their distilled variants (SD3.5-Large-Turbo and FLUX.1-Schnell), the norm of $\hat z_0$ consistently decreases as the timestep becomes smaller, reaching its minimum at $t=0$. For comparison, we include a theoretical curve derived from the Gaussian model by setting $|\mu|^2 = 512^2$ and $\sigma^2 = 0.03$, which approximates an image distribution concentrated on a high-dimensional hyperspherical shell.

For Fig.~\ref{fig:sup_fig_10} (a), although the empirical curves do not perfectly overlap with the theoretical one, they exhibit the same monotonic trend. The discrepancy mainly stems from the fact that the theoretical curve models the match between the marginal Gaussian noise distribution and the full image distribution. In contrast, real generative models operate at the instance level: during sampling, the model predicts the posterior mean of $z_0$ conditioned on a specific latent $\hat z_t$ and the text condition, rather than integrating over the full noise prior. As a result, the posterior $p(z_0 \mid z_t, c)$ becomes more concentrated and exhibits a nonzero conditional bias, causing the empirical $z_0$ norm–timestep curves to shift upward and slightly to the right relative to the idealized Gaussian prediction.

\begin{figure}[htb]
    \centering
    \includegraphics[width=1\linewidth]{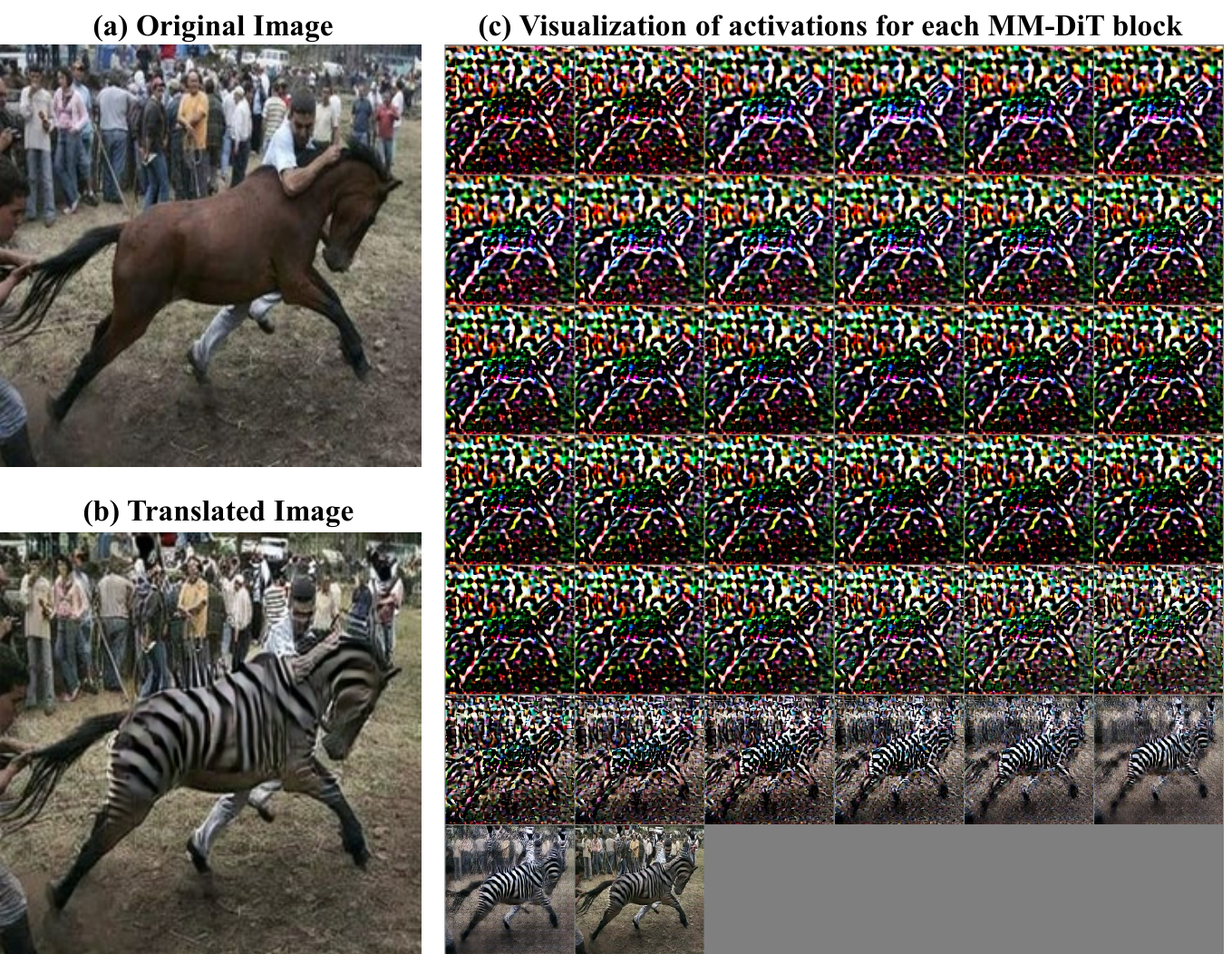}
    \caption{{\bf Visualizations of Activations Across MM-DiT Blocks.} 
    This experiment is conducted on SD3.5-Large-Turbo. The activations are shown in order from block 1 to block 38, arranged from left to right and top to bottom.
    }
    \label{fig:sup_fig_11}
\end{figure}

\section{Lightweight Modification for MM-DiT}
\label{sec:sup_lightweight_modification}

To improve inference speed, we implement lightweight modifications to MM-DiT by removing the text branch from the early MM-DiT blocks. As shown in Fig.~\ref{fig:sup_fig_11}, we visualize the activations after each MM-DiT block during inference. The first 30 activations change slightly, while the last 8 activations begin to transform from horse to zebra. We hypothesize that the earlier MM-DiT blocks mainly extract features from the input image, while the later blocks focus on modifying the image based on the text guidance. Therefore, the text branch in the early blocks is less critical and can be removed without significantly impacting model performance.

\section{Additional Results}
\label{sec:sup_addtional_vis_results}
In this section, we display more comparative results on 
 unpaired Horse2Zebra, BDD Day2Night, BDD Clear2Rainy, LHQ2Shinkai,
  and paired datasets Edge2Photo. The results are shown as follows.

\begin{figure*}[htb]
    \centering
    \includegraphics[width=0.95\linewidth]{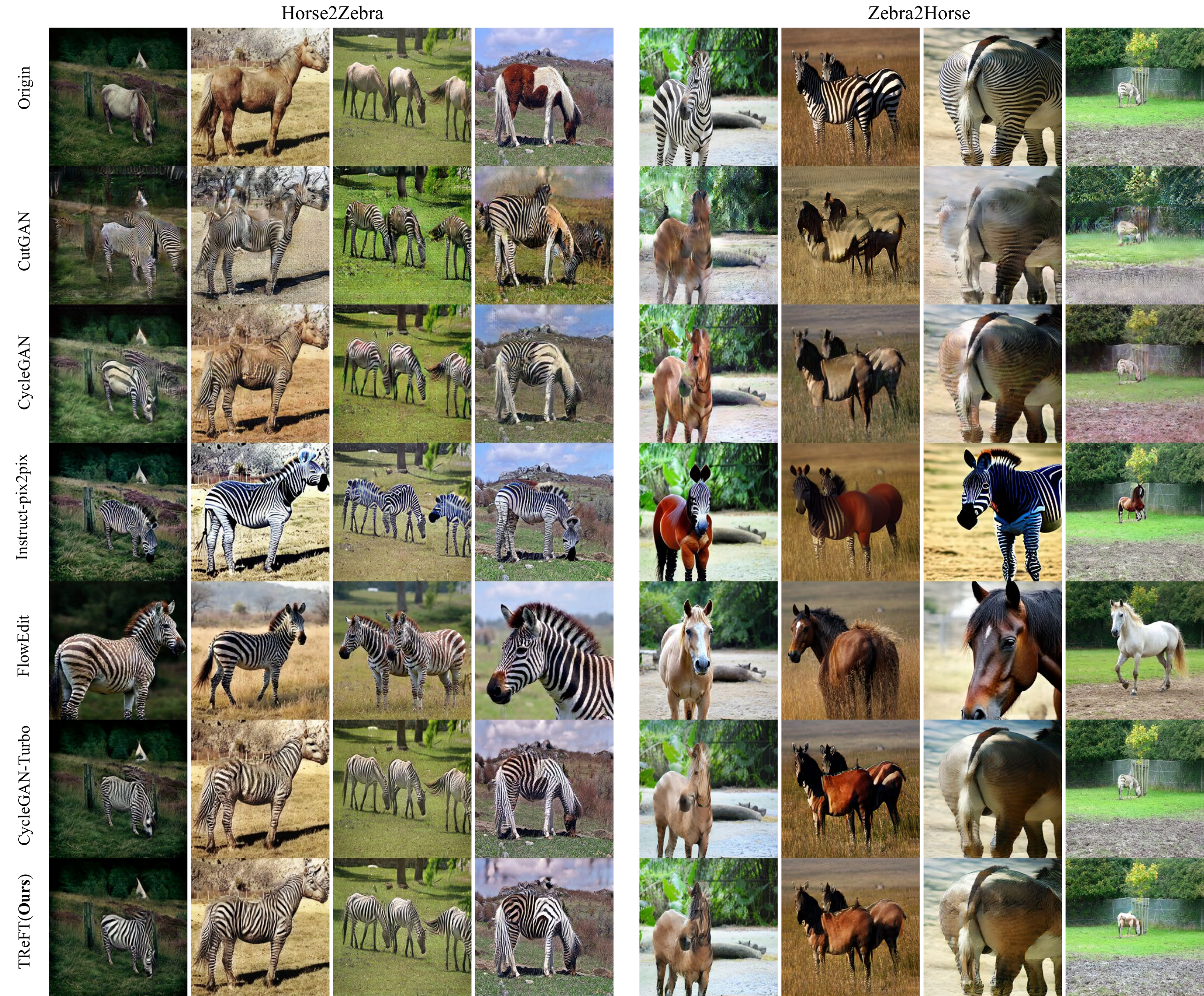}
    \caption{{\bf Additional comparative results.}}
\end{figure*}

\begin{figure*}[htb]
    \centering
    \includegraphics[width=0.95\linewidth]{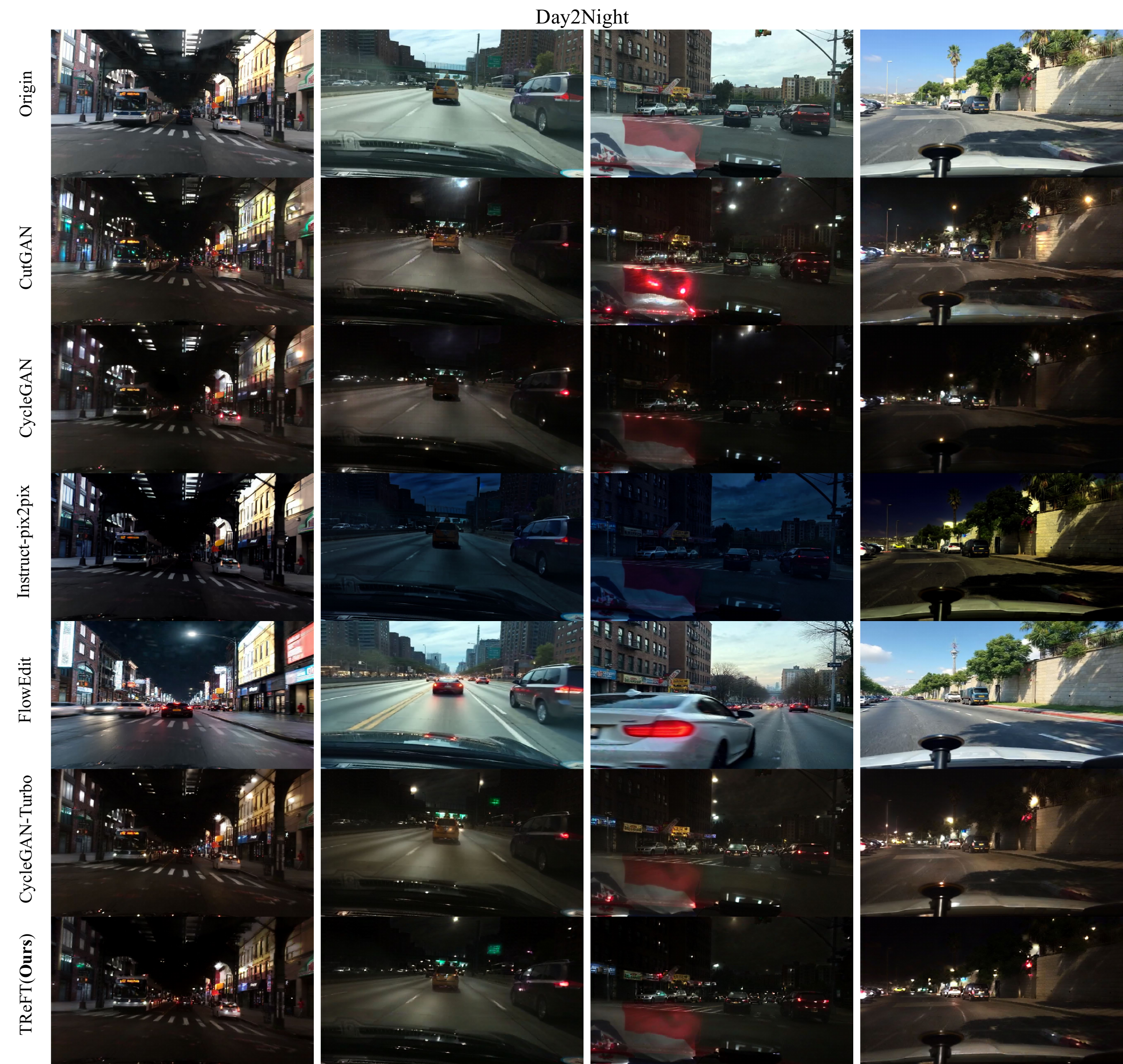}
    \caption{{\bf Additional comparative results.}}
\end{figure*}

\begin{figure*}[htb]
    \centering
    \includegraphics[width=0.95\linewidth]{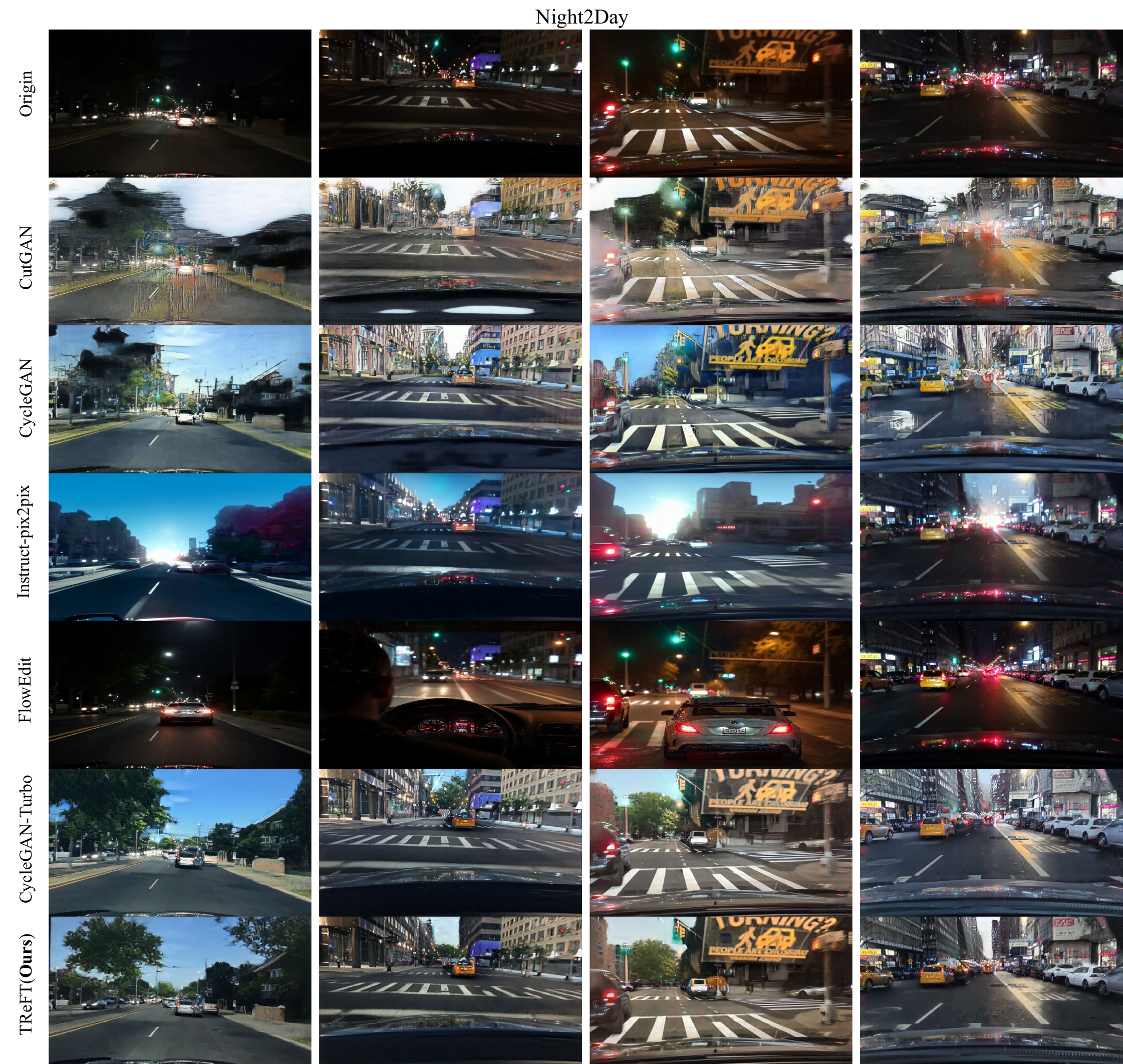}
    \caption{{\bf Additional comparative results.}}
\end{figure*}

\begin{figure*}[htb]
    \centering
    \includegraphics[width=0.95\linewidth]{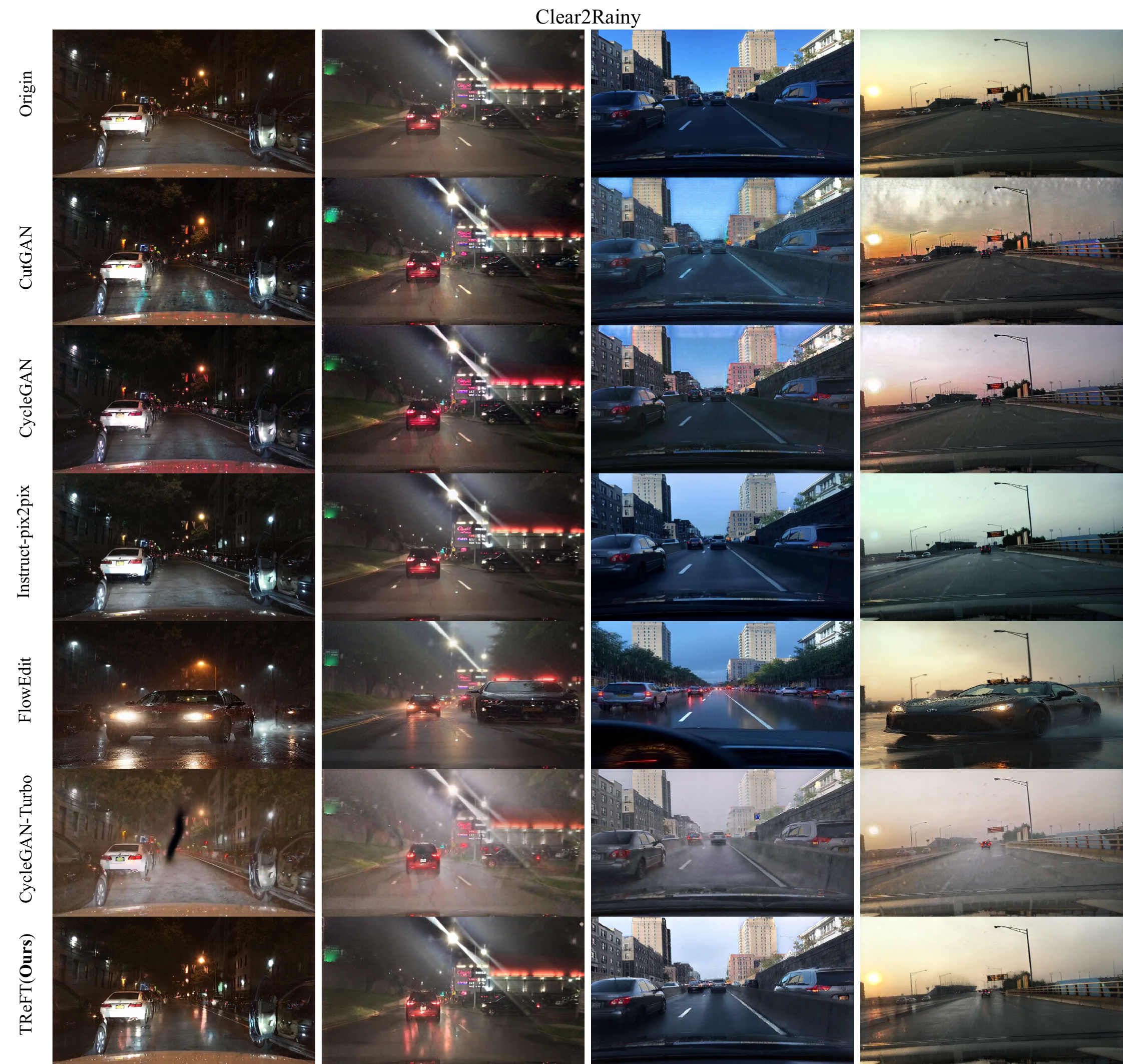}
    \caption{{\bf Additional comparative results.}}
\end{figure*}

\begin{figure*}[htb]
    \centering
    \includegraphics[width=0.95\linewidth]{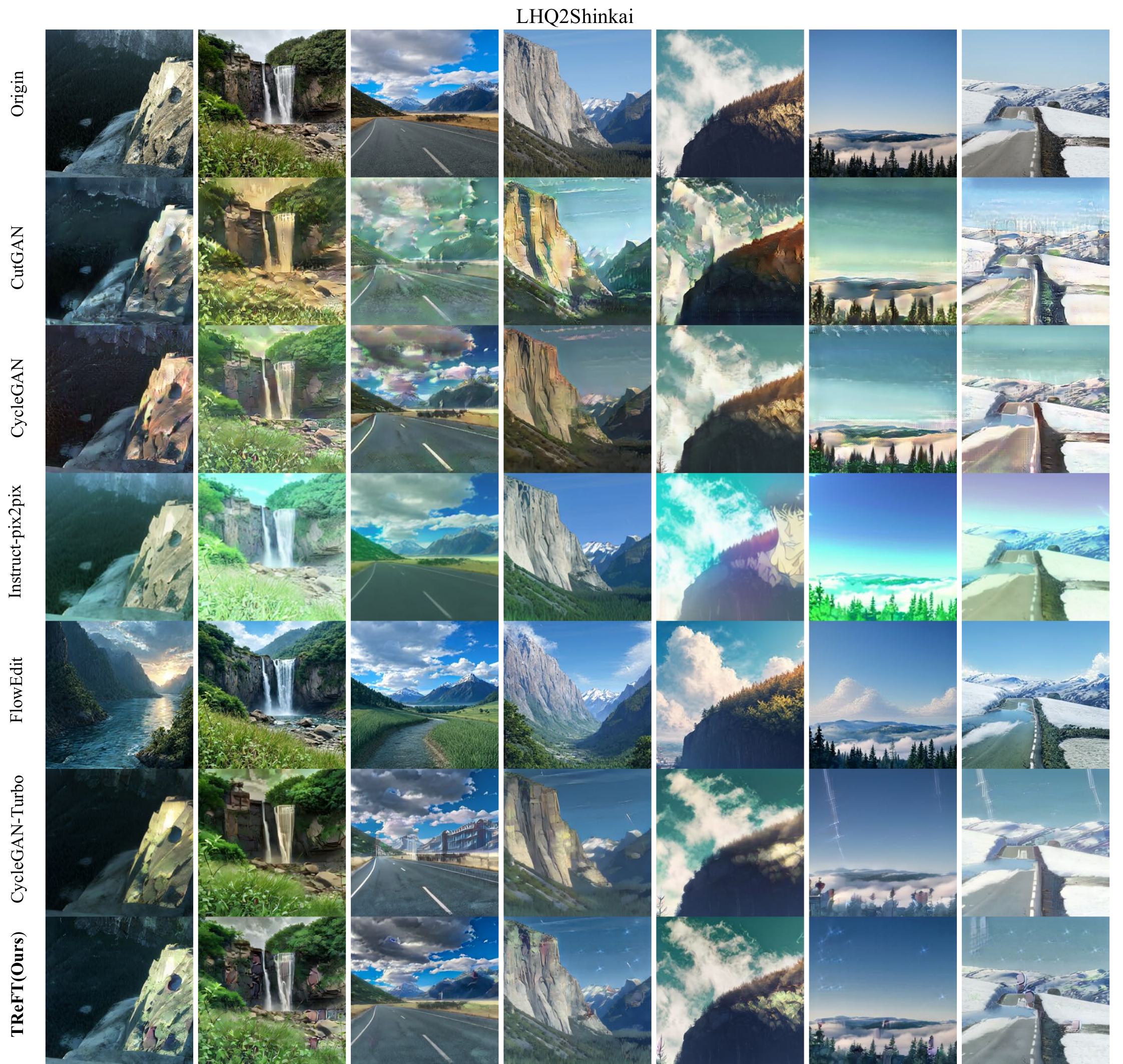}
    \caption{{\bf Additional comparative results.}}
\end{figure*}

\begin{figure*}[htb]
    \centering
    \includegraphics[width=0.55\linewidth]{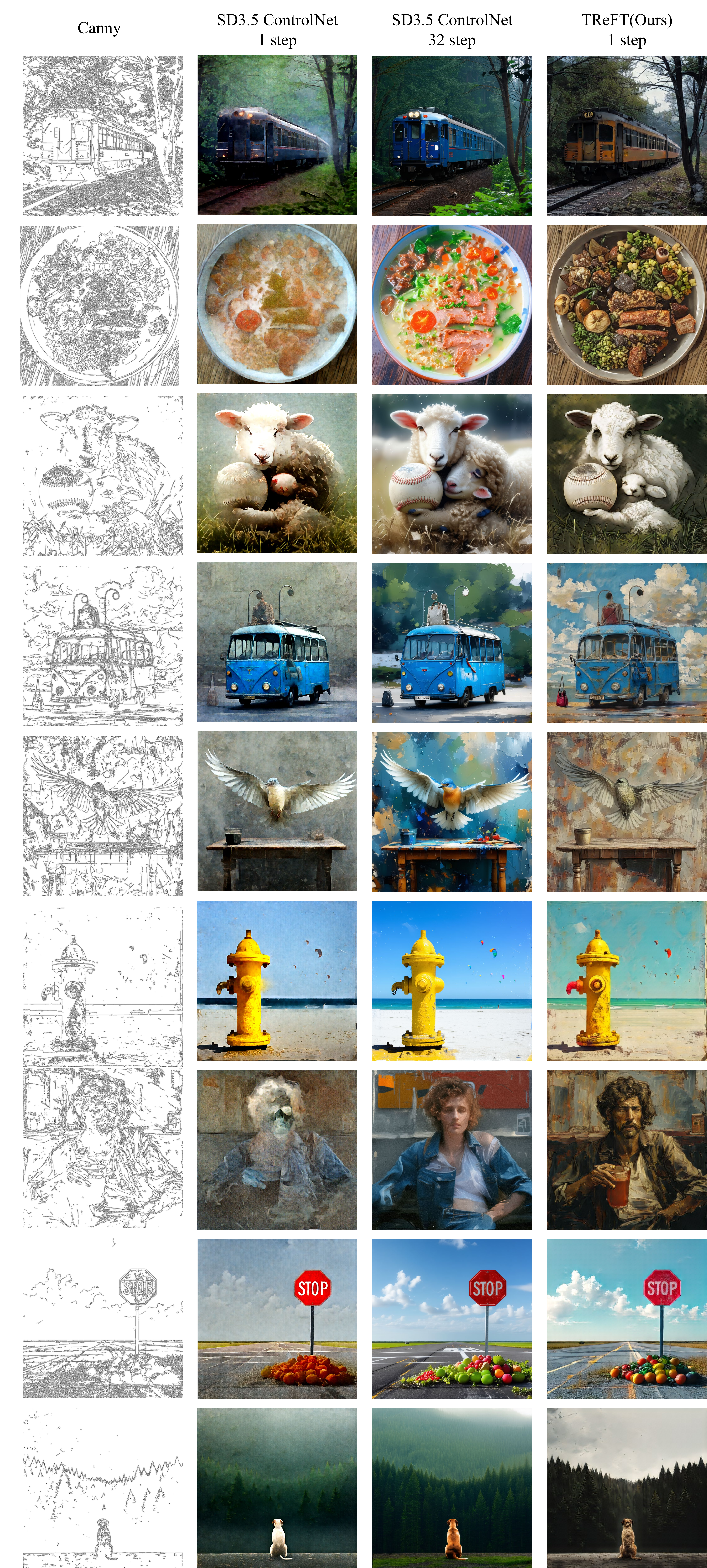}
    \caption{{\bf Additional comparative results.}}
\end{figure*}

\clearpage
\clearpage


\end{document}